\documentclass{article}

\usepackage[preprint]{tmlr}

\usepackage{amsmath,amsfonts,bm}

\def\eqref#1{equation~\ref{#1}}

\def\1{\bm{1}}

\DeclareMathAlphabet{\mathsfit}{\encodingdefault}{\sfdefault}{m}{sl}
\SetMathAlphabet{\mathsfit}{bold}{\encodingdefault}{\sfdefault}{bx}{n}

\usepackage{amsmath, amssymb, amsthm}
\usepackage{booktabs}
\usepackage{tabularx}
\usepackage{graphicx}
\usepackage{hyperref}
\usepackage{url}

\title{Testing Frontier Large Language Models' Physics Literacy\\
       in Parallel Physical Worlds}

\author{\name Dong Zhang \email dongzhanghz@gmail.com}

\def\month{MM}
\def\year{YYYY}
\def\openreview{\url{https://openreview.net/forum?id=XXXX}}

\begin{document}

\maketitle

\begin{abstract}
\noindent
Current large-language-model (LLM) physics benchmarks are usually scored by answer accuracy, which cannot distinguish genuine reasoning from recall of familiar problem patterns and reveals little about where a model's reasoning breaks down. We introduce an auditable four-stage diagnostic that evaluates whether an LLM can reason inside an unfamiliar physics framework through induction, formulation, prediction, and review. The diagnostic combines locked pre-registrations, fresh sessions between stages, dual-LLM judging, and a human-audit pathway, and we apply it to three parallel physics worlds: a single-equation counterfactual world ($F=mv$), a historical framework (Aristotelian mechanics), and a four-domain counterfactual world (Decay World). Across Claude Opus 4.7, GPT-5.5, and Gemini 3.1 Pro, the three worlds yield composite PASS rates are 6/15, 6/15, and 0/15 respectively (content $\land$ structural for $F=mv$ and Aristotelian, content axis only for Decay World where the structural axis is out of scope). The most pointed empirical pattern is a qualitative-versus-quantitative asymmetry: in Decay World, models almost never predict the wrong direction of change, but frequently compute the wrong ratio by slipping back to standard-physics relations. The protocol also surfaces two methodology findings: LLM-judge reliability does not transfer across frameworks, and Stage 4 self-review is weak in every framework, with the model's own review wrongly reporting no earlier error in at least two-thirds of the trials that actually contained one. We release the full prompts, responses, verdicts, and audit records.

\end{abstract}


\section{Introduction}\label{sec:intro}

Whether large language models (LLMs) can reason is one of the central debates in AI research. In the literature, reasoning is usually defined through tasks that cannot be answered in a single step, such as multi-step arithmetic, geometric proofs, and long-horizon planning, where the model must produce intermediate steps that can be checked independently. \citet{wei2022chain} introduced Chain-of-Thought (CoT) prompting in 2022: supplying a few ``question, intermediate steps, answer'' examples in the prompt raised PaLM-540B's accuracy on GSM8K~\citep{cobbe2021training} from 17.9\% to 56.9\%. Interestingly, Kojima et al. later found that examples were not even necessary: the single phrase ``Let's think step by step'' raised zero-shot InstructGPT-175B from 10.4\% to 40.7\%~\citep{kojima2022zero}. Together with \citet{brown2020language}'s GPT-3 scaling observation and \citet{wei2022emergent}'s emergence hypothesis, these results supported the optimistic view that scale plus prompting can produce emergent reasoning. On the other hand, this optimistic view has been contested by Schaeffer et al., who argued that much of the observed emergence is an artifact of the evaluation metric~\citep{schaeffer2023emergent}. Mirzadeh et al.'s GSM-Symbolic showed that frontier models are highly sensitive to symbolic perturbations of the problem text, with accuracy in the GSM-NoOp variant dropping by more than 60 percentage points~\citep{mirzadeh2024gsm}. This suggests that surface symbol matching contributes far more than true multi-step derivation. Huang and Chang's survey places this debate on a spectrum, from ``CoT is real reasoning'' to ``CoT is template retrieval from training data''~\citep{huang2023reasoning}. Whether LLMs perform genuine reasoning thus remains an open question.

A more ambitious question goes beyond reasoning: can AI do scientific research? We sort existing ``AI for science'' work into three levels of autonomy, all of which remain anchored to a framework supplied in advance. \textbf{Low autonomy}: AI acts as an execution tool within an established theory or experimental pipeline, carrying out optimization, automation, parameter search, or data analysis. Boiko et al.'s Coscientist (2023), which let an LLM autonomously schedule organic synthesis experiments, and the Finch analysis module in Robin are examples of this role~\citep{boiko2023autonomous, ghareeb2026robin}. \textbf{Medium autonomy}: AI generates and filters hypotheses across a large search space before handing candidates to humans or downstream experiments. FunSearch searched programs under a human-written evaluation function~\citep{romera2024mathematical}, Google DeepMind's Co-Scientist generated drug-repurposing hypotheses across the literature~\citep{gottweis2026coscientist}, Robin combined literature review, hypothesis evaluation, and data analysis in a closed loop for dry age-related macular degeneration~\citep{ghareeb2026robin}, and The AI Scientist ran variants inside a human-written machine-learning template~\citep{lu2024ai}. \textbf{High autonomy}: AI solves a precisely stated problem whose success criterion and external verification are supplied by humans, producing an independently checkable object inside a rigorously defined search space. AlphaGeometry produced formalized proofs of IMO geometry problems~\citep{trinh2024solving}, and OpenAI reported in May 2026 that an internal reasoning model gave a counterexample construction for Erd\H{o}s's 1946 unit distance conjecture, verified jointly with nine external mathematicians~\citep{openai2026erdos}.

These three levels share one structural feature: the problem being solved sits inside a theoretical framework that humans have established in advance. The AI's role is to search, combine, reason, or prove within that framework. The framework itself remains fixed throughout the solution. In other words, the three levels correspond to different fragments of the scientific research chain. Specifically, low autonomy corresponds to experimental execution and data processing within an existing framework. Medium autonomy corresponds to candidate hypothesis generation on top of known regularities. High autonomy corresponds to a solution or construction under a given problem statement. In all three cases, the AI completes one stage of theory construction, not the construction of a complete theory itself. A complete scientific theory requires starting from theoretically uninterpreted observations, independently inducing a theoretical system that a third party can check, and making quantifiable predictions about new situations from that system. This process is central to physics training, but, to our knowledge, no AI system has independently completed it.

Another research direction adjacent to complete scientific theory construction is world modeling, increasingly presented in industry as infrastructure for ``physical AI'' systems~\citep{nvidia2025cosmos}. However, the term ``world model'' has no unified definition in the literature. Ha and Schmidhuber's world model is a recurrent network that predicts the next environment state given the current state and action~\citep{ha2018world}. LeCun's JEPA proposal goes the other way: a predictive module in a latent representation space, not at the pixel level~\citep{lecun2022path}. Meanwhile, the physics question is more pointed. OpenAI initially described Sora as an implicit physics simulator\footnote{OpenAI shut down the Sora consumer product in April 2026 and plans to retire the API in September.}~\citep{brooks2024sora}, but LeCun argued that pixel-level generation and a causal model of the world are different objects~\citep{lecun2022path}. Bear et al.'s Physion benchmark (2021) measures next-step prediction of rolling, sliding, and colliding from visual input~\citep{bear2021physion}. This is forward physical prediction, and it does not require the model to externalize rules in an independently checkable form. World modeling has largely been pursued as an engineering problem. Even the line that overlaps with cognitive science, such as Battaglia, Hamrick, and Tenenbaum's intuitive physics engine, treats ``physics'' at the level of perception and intuition rather than the induction of a formal physical theory from observations~\citep{battaglia2013simulation}. Overall, world modeling provides a necessary but insufficient foundation. A model without an internal representation of the world cannot produce an independently checkable formal theory, but having that representation does not mean it can write the theory down. To assess physics understanding, a world model must be examined from the standpoint of physics itself.

To study how world models understand physics, let us return to the LLM. How does the literature currently assess LLM physics understanding? The dominant approach is to evaluate its problem-solving skills. GSM8K~\citep{cobbe2021training} uses grade-school math word problems, and MATH~\citep{hendrycks2021measuring} extends to high-school competition problems. Both are largely saturated by frontier models today. Rein et al.'s GPQA Diamond pushes difficulty up with 198 PhD-level questions across three science disciplines, where in-domain PhD experts score 65\% and Claude 3.5 Sonnet 59.4\%~\citep{rein2024gpqa}. Qiu et al.'s PHYBench uses 500 original physics problems and introduces an Expression Edit Distance (EED) score that catches cases where the process is correct but the answer is wrong~\citep{qiu2025phybench}. On straight accuracy, Gemini 2.5 Pro reaches 36.9\% and human experts 61.9\%. Zhang et al.'s ABench-Physics rewrites a problem as a sequence of dynamic variants and requires the model to get every variant right~\citep{zhang2025abench}. This is the same perturbation-as-fragility-test idea as Mirzadeh et al.'s GSM-Symbolic~\citep{mirzadeh2024gsm}.

The phenomena above point to the same structural gap. Perturbation tests on reasoning show that benchmark correctness can rely substantially on symbolic matching (\citet{mirzadeh2024gsm}). Existing AI-for-science systems pose problems inside theoretical frameworks that humans have already established, so the induction of the framework itself is not independently tested. World-model evaluations keep ``physics'' at the level of perception and next-step prediction, which is not the same as a formal theory in the physicist's sense~\citep{zhang2024sora}. LLM physics benchmarks, even after improvements such as EED scoring in PHYBench~\citep{qiu2025phybench} and dynamic variants in ABench-Physics~\citep{zhang2025abench}, share the same accuracy-as-criterion logic and inherit two structural limitations. First, accuracy cannot in principle distinguish physics understanding from prior exposure to similar problems. A bigger benchmark with more problems naturally yields more correct answers without implying more cognitive ability. Second, accuracy carries no information about a model's cognitive boundary. If model A scores 90\% and model B scores 91\%, the reader learns only that B got one more problem right, not what kind of cognitive work B can do that A cannot. Closing this gap requires moving beyond problem-solving and directly observing whether a model can start from unfamiliar phenomena and independently produce a theoretical system that a third party can check.

The history of physics contains a reproducible cognitive chain. For example, Kepler induced three laws from twenty years of Tycho Brahe's planetary position data: elliptical orbits, equal areas, and $T^2 \propto a^3$~\citep{kepler1609astronomia}. This was pure geometry and kinematics, without the concept of ``force.'' Newton, starting from Kepler's geometric laws and Galileo's falling-body observations, induced universal gravitation and used it to predict tides, comet orbits, and lunar motion~\citep{newton1687principia}. Maxwell, starting from Coulomb's electrostatics, Ampere's circuital law, Faraday's electromagnetic induction, and Gauss's laws of electricity and magnetism, recognized that Ampere's law was missing a term required for self-consistency~\citep{maxwell1865dynamical}. He added the displacement current, unified the phenomena into four equations, and used those equations to predict electromagnetic waves propagating at the speed of light, implying that light itself is an electromagnetic wave. Einstein started from the constancy of the speed of light (Michelson-Morley), Maxwell electrodynamics, the equivalence principle (free fall as a local inertial frame), and the anomaly in Mercury's perihelion precession under Newtonian gravity~\citep{einstein1916grundlage}. He built special and general relativity and predicted the bending of light and the existence of gravitational waves. The same cognitive chain appears repeatedly in the history of complete theoretical systems in physics: observe phenomena, induce regularities, systematize them into operational form, predict new situations, and then review the theory to identify errors and boundaries. Physics is physics precisely because it often requires walking this whole chain. DeepMind CEO Hassabis proposed in early 2026 an AGI test: give an AI system all knowledge prior to 1911 and see whether it can independently produce Einstein's 1915 general relativity~\citep{hassabis2026einstein}. The merits of that specific test are beyond the scope of this paper, but its premise is clear. An AI system would have to walk the full chain of observation, induction, systematization, prediction, and review.

Given this, our test of whether an LLM understands physics decomposes that cognitive chain into four fundamental cognitive moves and evaluates each separately. To keep the four moves concrete, we trace one running example through all of them, the development of Maxwell's electromagnetic theory:
\begin{itemize}
  \item \textbf{Induction}: extract from theoretically uninterpreted observations a regularity that explains all of them at once. Maxwell started from four seemingly independent groups of electric and magnetic phenomena accumulated in the 19th century (Coulomb's electrostatics, Ampere's circuital law, Faraday's electromagnetic induction, and Gauss's laws of electricity and magnetism) and induced that they could all be described by a single underlying structure.
  \item \textbf{Formulation}: rewrite the induced regularity into an operational form that a third party can apply. Maxwell systematized this induction into four partial differential equations (Maxwell's equations) and, in doing so, added the displacement current term to preserve the equations' internal self-consistency.
  \item \textbf{Prediction}: use the operational form to make quantitatively checkable predictions about new situations. Maxwell's equations directly imply electromagnetic waves and give their propagation speed as the measured speed of light $c$, predicting that light is an electromagnetic wave. Hertz confirmed this prediction in 1888.
  \item \textbf{Review}: look back at the theory one has built and identify which step might be wrong, which boundary has not yet been covered, and which mechanism has not yet been explained. Reviewing Maxwell's equations reveals that the equations require the electromagnetic wave speed $c$ to be the same in all inertial frames, while Galilean transformations cannot preserve their covariance. To keep the system self-consistent, Galilean transformations must be replaced by Lorentz transformations. Maxwell's system already contains the seed of special relativity.
\end{itemize}
Note that these four cognitive moves are not borrowed from a philosophy-of-science framework centered on ``scientific revolutions.'' Popper argued that scientificness is defined by falsifiability, and that theories advance through bold conjectures combined with rigorous attempts at refutation~\citep{popper1959logic}. Kuhn described science as an alternation between normal science and paradigm shifts, focusing on when an old paradigm is replaced by an incommensurable new one~\citep{kuhn1962structure}. Both are concerned with when theories are overturned and paradigms are replaced. We draw instead on an older tradition of scientific method: Whewell's account of induction, colligation, prediction, and consilience, and Hempel's later formal treatment of scientific explanation in the hypothetico-deductive and deductive-nomological tradition~\citep{whewell1840philosophy,hempel1966philosophy}. In this tradition, a physicist in routine work starts from observations and successively performs induction, formulation, prediction, and review to build a theoretical system that a third party can apply. This is the working path that repeatedly appears in physics training, not a description of revolutionary moments.

However, running this evaluation inside a real physics framework creates a fundamental difficulty. Frontier LLMs have absorbed huge volumes of physics textbooks, exam problems, and research papers during training~\citep{soldaini2024dolma,grattafiori2024llama3}. Standard conclusions from real physical laws have appeared repeatedly in their training corpora. When a model performs well on these frameworks, we cannot tell whether it is genuinely performing induction, formulation, prediction, and review, or simply reproducing conclusions it has memorized during training~\citep{carlini2023quantifying,sainz2023nlp,zhang2024careful}. An LLM is a black box whose outputs are tightly coupled to training data. This is the root reason why the LLM physics benchmarks discussed above, even with EED scoring or dynamic variants, cannot answer the question of whether the model does physics.

Our response is to move the test into counterfactual physics frameworks, following the broader observation that counterfactual task variants can separate reasoning from recitation~\citep{wu2024reasoning}. We posit a set of ``parallel physics worlds'' whose laws differ from real physics. We describe their phenomena in everyday language and deliberately exclude modern physics terminology. In these frameworks, the model cannot have seen a standard answer during training. To walk through induction, formulation, prediction, and review, it must start from the given phenomena rather than from physical rules already memorized from training data. Each framework consists of an observation set, a set of criteria, and a set of application scenarios. The model's output at each step is judged independently, and only when every required step passes do we count the run as a composite PASS. When a model forces real physical rules onto counterfactual phenomena, the failure mode itself becomes direct evidence about its physical-reasoning capacity. In this paper, we conduct experiments in three parallel physics worlds that form a graded difficulty progression, and test several current frontier LLMs. Concurrent work probes LLM rule discovery in simulated physics worlds with non-canonical laws~\citep{wiemann2026discoverphysics,zheng2026newtonbench}. Our work differs along three axes. First, it tests the full four-move cognitive chain, including the final review move that asks whether the model identifies the errors in its own derivation. Second, it presents observations in everyday language with modern physics terminology removed, rather than as numerical trajectories or symbolic forms. Third, it includes a historical framework (Aristotelian mechanics) alongside two counterfactual ones, probing whether the model can suspend the criticisms attached to a theory in its training data. Section~\ref{sec:methods} gives the details of the three worlds, the model list, the evaluation protocol, and the judgment criteria.

Building on this, our contributions are organized in three layers, each of which is, to our knowledge, the first systematic instance of its kind in the LLM physics-evaluation literature.
\begin{itemize}
  \item \textbf{Methodology layer}: we operationalize the four cognitive moves into four independent evaluation tests, each judged separately under explicit criteria, with a composite-PASS gate at the end. The closest prior work~\citep{wiemann2026discoverphysics,zheng2026newtonbench} treats physics reasoning as a single end-to-end law-discovery process. To our knowledge, this is the first decomposition of physics reasoning into the four moves of this chain measured as independent axes, including the Review move that asks whether the model identifies the errors in its own derivation.
  \item \textbf{Experimental framework layer}: we design three parallel physics worlds that exert systematically different cognitive pressures: a single-equation counterfactual ($F=mv$), a historical framework (Aristotelian mechanics) probing whether the model can suspend the criticisms attached to a coherent but training-data-disfavored theory, and a four-domain counterfactual (Decay World) without an underlying substrate. No existing physics-reasoning benchmark, as far as we know, pairs historical and counterfactual frameworks within a single graded difficulty progression.
  \item \textbf{Openness and auditability layer}: the evaluation pipeline is open end to end, from prediction lock-in through multiple LLM judges to human audit. All raw prompts, model responses, judge outputs, and human-audit records are independently reproducible. It is also, we believe, the first publicly available end-to-end auditable pipeline for LLM physics evaluation, contributing one set of reproducible data points to the LLM-as-judge literature and another to the LLM physics-evaluation literature.
\end{itemize}

\section{Methodology}\label{sec:methods}
This section presents the full evaluation pipeline: how the four-stage protocol operationalizes the four fundamental cognitive moves into four independent tests, how pre-registration is engineered to be tamper-evident, how judgment is backstopped by multiple LLM judges plus human audit, and how the three ``parallel physics worlds'' frameworks are sourced and tiered.

\subsection{Four-Stage Protocol}\label{sec:methods:protocol}

The four cognitive moves identified in \S\ref{sec:intro} (induction, formulation, prediction, review) could in principle be tested in a single end-to-end run: hand the model an observation set and ask for a full theoretical account. We do not do this for two reasons. First, a single end-to-end test produces a single binary signal that mixes four different capabilities and hides which one is failing. A model strong at induction but weak at quantitative prediction is indistinguishable from a model weak at induction but strong at prediction. Second, a single end-to-end test invites carry-over errors: a model that gives a flawed inductive rule in the first move will reuse it downstream, conflating ``failed at induction'' with ``failed because it inherited a bad induction''. Splitting the chain into four independent stages breaks both confounds.

\begin{figure}[h]
\centering
\includegraphics[width=\textwidth]{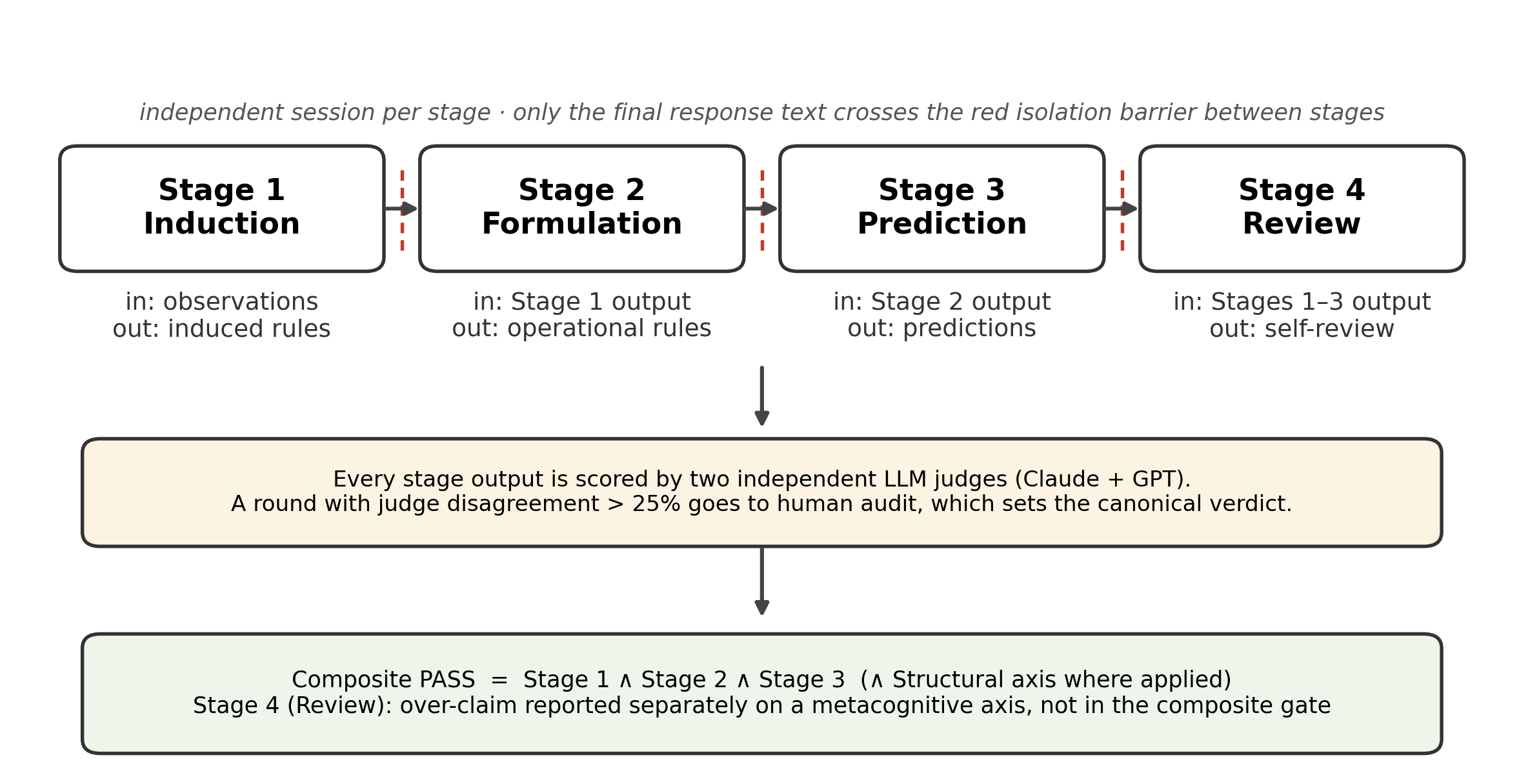}
\caption{Four-stage protocol and evaluation pipeline. A trial runs one tested model through Stage 1 (induction), Stage 2 (formulation), Stage 3 (prediction), and Stage 4 (review), with each stage in an independent API session and only the final response text of Stage $k$ passed to Stage $k+1$. Every stage output is scored by two independent LLM judges (Claude and GPT). If a round's judge-disagreement rate exceeds 25\%, the disagreement cases go to human audit for the canonical verdict (\S\ref{sec:methods:judge}). The composite verdict requires Stage 1, Stage 2, and every Stage 3 scenario to pass, plus the structural axis where the framework applies one (\S\ref{sec:methods:pass}). Stage 4 over-claim is reported separately as a metacognitive axis. Each framework is run on Claude Opus 4.7, GPT-5.5, and Gemini 3.1 Pro with $N=5$ trials per model. Gemini is tested but not used as a judge.}
\label{fig:four-stage-protocol}
\end{figure}

We test the four moves in sequence (see Figure~\ref{fig:four-stage-protocol}). Each stage receives the previous stage's final text output as its input and produces an output that becomes the input to the next stage:

\begin{itemize}
  \item \textbf{Stage 1 (Induction)}: input is the framework's observation set, a set of 10--12 natural-language descriptions of phenomena in the parallel physics world. Output is a set of induced rules.
  \item \textbf{Stage 2 (Formulation)}: input is the model's own Stage 1 induction. Output is the rules rewritten in operational form, with explicit definitions, scopes of applicability, and boundary cases.
  \item \textbf{Stage 3 (Prediction)}: input is the model's own Stage 2 operational rules. Output is a set of quantitative predictions on a set of unseen application scenarios.
  \item \textbf{Stage 4 (Review)}: input is the model's own outputs from Stages 1 through 3. Output is a self-assessment that identifies which earlier step, if any, contains an error.
\end{itemize}
A point worth stressing: the model at Stage $k+1$ is conversing with its own past output, not with the experimenter. There is no human in the loop between stages.

To make each stage's output reflect the independent capability of that move, we enforce strict isolation between stages. Each stage runs in a fresh API session, with a new client and a new session UUID, and the protocol requests temperature 0 where the provider exposes the parameter. Cross-stage context reuse is forbidden. The model at Stage $k+1$ receives only the final response text of Stage $k$. It does not see the chain-of-thought, the reasoning trace, or any intermediate state from Stage $k$. This design rules out a class of confounds in which the model would otherwise carry hidden internal state forward and pass off one stage's reasoning as another stage's capability.

Each stage has its own type of criterion, recorded in the framework's pre-registration file. Stage 1 uses a banned-word list, a set of necessary conditions, a set of suspicious failure modes, and a halt-at-first-FAIL procedure. Stage 2 uses operational-form criteria that check whether each rule has an operational definition, a stated scope of applicability, and a treatment of boundary cases. Stage 3 uses an explicit numerical PASS interval for each scenario. Stage 4 uses a three-category classification: correct self-identification, missed error, or vacuous over-claim. The full criteria for each framework are given in Sections~\ref{sec:fmv}--\ref{sec:decay}.

The basic unit of evaluation is a trial. One trial is one model running through all four stages on one framework's observation set. For each framework we run 3 models $\times$ $N=5$ trials, for a total of 15 trials. The composite verdict for a trial requires that Stage 1, Stage 2, and every Stage 3 scenario pass independently, and, where the framework applies a structural axis, that the structural axis also passes (see \S\ref{sec:methods:pass}). Stage 4 is reported as a separate metacognitive axis and does not enter the composite gate.

\subsection{Models and Run Configuration}\label{sec:methods:config}

We test three frontier large language models: Claude Opus 4.7, GPT-5.5, and Gemini 3.1 Pro. Each is pinned to the exact provider version string in effect for its round, recorded in the committed run artifacts. Appendix~\ref{app:prereg-tags} identifies the locked pre-registration tags. After every API call the runner verifies the returned model identifier against the pinned string and aborts on any mismatch. For each framework we run $N=5$ trials per model, for 15 trials per framework. The protocol requests temperature 0. Where a provider does not expose the parameter, the runner records the requested value and the call uses the provider's default sampling. Every stage runs in a fresh API session with a new client and session UUID and no context reuse across stages (\S\ref{sec:methods:protocol}). Pass/fail judging at every stage is performed independently by two LLM judges from different providers (Claude and GPT), detailed in \S\ref{sec:methods:judge}.

\subsection{Irreversible Pre-Registration}\label{sec:methods:prereg}

A standard hazard in evaluation work is post-hoc rationalization: once a judgment criterion produces an inconvenient result, the criterion can quietly be relaxed until the result becomes convenient. The danger is acute in LLM evaluation because most criteria are written in natural language and are therefore easy to reword~\citep{sculley2018winners}. To make this kind of drift impossible rather than merely discouraged, we lock the full evaluation specification before any production data exists~\citep{pineau2021improving}.

For each framework we produce a single pre-registration~\citep{nosek2015promoting} file before any production data is generated. The file contains the framework's pre-registered predictions, every stage's pass/fail criteria, the verbatim prompts sent to the tested model, and the scripts that judge the responses. The framework-specific pre-registration files are described in their own chapters (Sections~\ref{sec:fmv}--\ref{sec:decay}).

The lock has two parts. A SHA-256 hash of the file's content is written into the file's own header, and a git tag is created at the corresponding commit. A pre-commit hook plus a CI check recomputes the SHA-256 on every commit and pull request and compares it to the value in the file header. Any silent edit, even a one-character change, makes the two hashes diverge and the commit fails. The pre-registered evaluation specification is therefore an engineering artifact rather than a promise: tampering with it would have to be explicit and visible in the git history.

If we find a genuine flaw in a pre-registered prediction or criterion after running experiments, we do not edit the existing pre-registration file. We create a new version with its own tag and SHA-256, and any results published under the new version explicitly flag the deviation. The original pre-registration remains in place as the canonical reference for the experiments that were run under it.

\subsection{Dual LLM Judges with Human Audit}\label{sec:methods:judge}

The PASS/FAIL judgments at each stage depend on semantic criteria: for instance, whether an induced rule explains all observations, whether an operational form states its scope of applicability, whether a self-assessment correctly identifies an earlier error. Judging hundreds of such outputs by hand is not practical, but a single LLM judge introduces a single point of bias and is itself an evaluation artifact. We use a dual-LLM-judge plus threshold-triggered human-audit scheme that combines scale with rigor.

Every stage output is judged independently by two frontier LLM judges from different providers (Claude and GPT, with exact version strings pinned per round). The two judges run on the same input under the same prompt, against the same pre-registered criteria, but in independent API sessions. Their verdicts are recorded verbatim alongside the model responses.

For each round, we measure the inter-judge disagreement rate across all dual-judged outputs. If the disagreement rate for any framework exceeds 25\%, a human audit is triggered before the round's results are released. The audit reviews every case where the two LLM judges disagreed, produces a canonical verdict, and the canonical verdict replaces the LLM verdicts for those cases in the final reported composite PASS. The 25\% threshold itself is pre-registered. We chose 25\% as a value high enough to absorb the random disagreements that two LLM judges produce on borderline cases, and low enough that any systematic semantic gap between the two judges forces the case to a human.

All raw judge outputs and the human-audit log when triggered are committed to the repository alongside the model responses (see the Code and Data Availability statement at the end of the paper).

\subsection{Defining the Composite Verdict}\label{sec:methods:pass}

The four cognitive moves are interdependent. A correctly induced rule that the model cannot apply consistently is not a usable theory. Conversely, correct predictions from an incoherent operational form are not theory construction. We therefore require all relevant stages to pass before counting a trial as a success.

The base composite verdict for a trial $t$ is the content-axis conjunction: Stage 1 PASS, Stage 2 PASS, and every Stage 3 application scenario PASS independently. Formally,
\[
  \text{composite}_{\mathrm{content}}(t) = \text{Stage1\_PASS}(t) \wedge \text{Stage2\_PASS}(t) \wedge \bigwedge_{s \in S} \text{Stage3\_PASS}(t, s),
\]
where $S$ is the framework's pre-registered set of application scenarios. A single Stage 3 scenario failure is sufficient to fail the entire trial. For frameworks that introduce an additional judging axis on top of the content axis, that axis joins the conjunction. We use a structural axis on the rule set (defined in \S\ref{sec:fmv:intro}) for $F=mv$ and Aristotelian, so the composite verdict reported in Sections \ref{sec:fmv} and \ref{sec:aristotelian} is $\text{composite}_{\mathrm{content}} \wedge \text{Structural\_PASS}$. Decay World (\S\ref{sec:decay}) does not apply the structural axis and its composite verdict is the content-axis composite alone. The headline composite PASS rates reported throughout the paper therefore have a per-framework composition stated explicitly in each framework's results section. We do not report partial composite scores or weighted averages, because doing so would obscure whether the model actually walked the full chain end to end.

Stage 4 (Review) is the model's self-assessment of its own earlier outputs. Including it in the composite gate would conflate two distinct capabilities: doing physics correctly and knowing when one has not. A model that produced a clean Stage 1--3 chain could still fail Stage 4 by missing a hypothetical error it was asked to identify. A model that produced a flawed Stage 1--3 chain could pass Stage 4 by correctly flagging its own errors. We therefore report Stage 4 separately as a metacognitive axis: for each framework we report the over-claim rate, defined as the fraction of failure-containing trials (those with at least one Stage 1--3 FAIL) in which the model's Stage 4 self-assessment claims that no earlier error occurred.

\subsection{Overview of the Three Parallel Physics Worlds}\label{sec:methods:worlds}

We run experiments in three parallel physics worlds (two counterfactual and one historical). The three are intended to place increasing cognitive pressure on the model, from shallow to deep, as a design hypothesis whose empirical signature is the bottleneck migration described in \S\ref{sec:cross-framework:gradient}. Table~\ref{tab:three-worlds} gives a side-by-side summary, and the three paragraphs below describe each world in turn.

\paragraph{$F=mv$ world (Easy).} The first world modifies a single equation: Newton's second law is replaced by $F=mv$, so that force is proportional to velocity rather than acceleration. All phenomena stay inside a single classical-mechanics domain. This is the shallowest level of the gradient, a single-equation counterfactual that still admits standard intuitions about force, mass, and motion.

\paragraph{Aristotelian mechanics (Medium).} The second world imports an entire historical framework. A body's natural motion is determined by its elemental composition (earth, water, air, fire), and an object moves toward where its dominant element naturally belongs. The framework is real, internally self-consistent, and widely covered by modern physics textbooks, almost always labeled as a ``counterexample.'' The cognitive pressure is not that the framework is unfamiliar but that the model has seen it many times in training, always with the label ``this is wrong.'' Reasoning inside the framework requires suspending that label.

\paragraph{Decay World (Hard).} The third world is a single rule: every directly measurable physical quantity loses 1\% of its value per second. The rule runs simultaneously across mechanical, thermal, rotational, and orbital domains. There is no underlying substrate (such as ``energy'') serving as the carrier of the rule, and all standard dissipation mechanisms (friction, damping, air resistance, viscosity, radiation) are explicitly turned off in the observation design. The cognitive pressure is multi-axis: a counterfactual rule, cross-domain unification of a single rate, and the absence of any conserved substrate.

\begin{table}[h]
\centering
\caption{The three parallel physics worlds used in this paper.}
\label{tab:three-worlds}
\begin{tabular}{llll}
\toprule
\textbf{World} & \textbf{Difficulty} & \textbf{Type} & \textbf{Core rule} \\
\midrule
$F=mv$        & Easy   & Counterfactual & $F=mv$ instead of $F=ma$ \\
Aristotelian  & Medium & Historical     & Element-determined natural motion \\
Decay World   & Hard   & Counterfactual & $1\%$/s decay across 4 domains \\
\bottomrule
\end{tabular}
\end{table}

The detailed design, observation sets, criteria, and pre-registration tags of the three worlds are given in Sections~\ref{sec:fmv} through~\ref{sec:decay}.

\section{Easy: the $F=mv$ Counterfactual World}\label{sec:fmv}

\subsection{Framework and observations}\label{sec:fmv:intro}

The $F=mv$ rule's numerical predictions diverge from $F=ma$ in obvious ways: a steady push gives steady speed rather than acceleration, a fall grows as $s = vt$ rather than $s = \tfrac{1}{2} g t^2$, and an object released from a moving hand drops straight down rather than along a parabola.

\textbf{The rule $F=mv$ never appears in any prompt the model sees.} What the model receives is 12 hand-written observations describing what a careful eye-witness would see in this world, with no underlying rule stated and no algebra. A sample observation reads, verbatim from Appendix~\ref{app:fmv-details}: \emph{The instant the steady pull begins, the block is already moving at its full steady pace. There is no gradual speeding-up from rest.} The phrasing is deliberately plain. Alongside the observations, a banned-word list keeps the model from importing the answer through vocabulary it has memorized: \emph{velocity}, \emph{acceleration}, \emph{momentum}, \emph{mass}, \emph{inertia}, \emph{gravity}, \emph{friction}, \emph{energy}, the equation $F = ma$ in any notation, and any physicist's proper name (\emph{Newton}, \emph{Newtonian}, \emph{Galileo}, \ldots). The test is purely lexical, applied to the whole response, including inflected forms. Naming a banned concept only to deny that it applies still counts as use.

So $F=mv$ is what the model has to find on its own. The Stage 1 instruction asks it to propose a self-consistent set of rules that explains every observation, using only the descriptive vocabulary the observations themselves use, and to return the rules as a numbered list. The instruction also asks the model to use as few rules as possible and to mark explicitly whenever one rule can be derived from another, so the rules form an axiomatic system rather than a flat list. If we had instead told the model the rule up front and asked it to predict what happens, we would be measuring a different capability. That kind of test is \textbf{deduction} from a counterfactual rule. Ours is \textbf{induction} of one.

What ``equivalent to $F=mv$'' means in practice is six necessary conditions N1--N6 on the induced rules, the \emph{content axis}. These are six criteria the judge applies, not a target number of rules: the model may induce as many or as few rules as it likes, and what matters is whether its rules jointly satisfy all six. N1 fixes the present-push principle: a body's pace at any moment is set by the push acting on it at that moment. N2 and N3 fix the two proportionalities: greater push gives greater pace for a given body, and greater heaviness gives smaller pace under the same push. N4 fixes the no-build-up, no-carry-over signature: a body is at its full pace as soon as the push acts, and stops the instant the push ends. N5 fixes the unified-fall picture: heavy and light fall alike at one unchanging pace. N6 fixes push combination: same-direction pushes add, opposite-direction pushes subtract. Failing any one of N1--N6 is a Stage 1 FAIL. A separate list of disqualifying surface patterns F1--F7 (\emph{e.g.}\ a rule stating that a push ``builds up'' pace, or that a body ``coasts'' after the push ends) gives further automatic FAIL triggers. Both N1--N6 and F1--F7 in full are in Appendix~\ref{app:fmv-details}.

A second, orthogonal \emph{structural axis} judges not what each rule says but how the rule set as a whole hangs together. Four conditions N9--N12 cover parsimony (the rule count should not vastly exceed the 12 observations), independence (no two Stage 1 rules paraphrase the same claim), traceability (every rule maps to specific observations, no fabricated mechanism), and hierarchy (a rule set of five or more should include cross-rule references, not be a flat enumeration). Full thresholds and FAIL triggers are in Appendix~\ref{app:fmv-details}. A trial's composite verdict in this section is content $\land$ structural: it PASSes only if it PASSes Stage 1, Stage 2, and Stage 3 on the content axis and PASSes the structural axis.

\subsection{Results}\label{sec:fmv:results}

We ran $N=5$ trials per model, for 15 trials in total. Table~\ref{tab:fmv-trials} gives the post-audit per-trial classification. The Over-claim column reports the Stage 4 metacognitive axis and is discussed for all three frameworks together in \S\ref{sec:cross-framework:overclaim}.

\begin{table}[h]
\centering
\caption{Post-audit per-trial classification for the $F=mv$ framework. $N=5$ trials per model, 15 trials total. Stage 1, Stage 2, Stage 3 are the content-axis stages judged on N1--N6. Structural is the rule-set axis judged on N9--N12 (Appendix~\ref{app:fmv-details}). Composite is content $\land$ structural: a trial PASSes only if Stage 1, Stage 2, Stage 3, and Structural all PASS. Over-claim is the separate Stage 4 metacognitive axis (\S\ref{sec:cross-framework:overclaim}): \texttt{yes}/\texttt{no} which is only defined on the content-FAIL trials.\\}
\label{tab:fmv-trials}
\small
\begin{tabular}{llcccccc}
\toprule
\textbf{Model} & \textbf{Trial} & \textbf{Stage 1} & \textbf{Stage 2} & \textbf{Stage 3} & \textbf{Structural} & \textbf{Composite} & \textbf{Over-claim} \\
\midrule
Claude Opus 4.7 & 0 & PASS & PASS & PASS & PASS & \textbf{PASS} & -- \\
Claude Opus 4.7 & 1 & PASS & FAIL & PASS & PASS & \textbf{FAIL} & no \\
Claude Opus 4.7 & 2 & PASS & FAIL & PASS & PASS & \textbf{FAIL} & yes \\
Claude Opus 4.7 & 3 & PASS & PASS & PASS & PASS & \textbf{PASS} & -- \\
Claude Opus 4.7 & 4 & PASS & PASS & PASS & PASS & \textbf{PASS} & -- \\
\midrule
GPT-5.5 & 0 & PASS & PASS & PASS & PASS & \textbf{PASS} & -- \\
GPT-5.5 & 1 & PASS & PASS & PASS & PASS & \textbf{PASS} & -- \\
GPT-5.5 & 2 & PASS & PASS & PASS & FAIL & \textbf{FAIL} & -- \\
GPT-5.5 & 3 & PASS & PASS & PASS & FAIL & \textbf{FAIL} & -- \\
GPT-5.5 & 4 & PASS & PASS & PASS & FAIL & \textbf{FAIL} & -- \\
\midrule
Gemini 3.1 Pro & 0 & FAIL & FAIL & PASS & PASS & \textbf{FAIL} & yes \\
Gemini 3.1 Pro & 1 & FAIL & FAIL & FAIL & PASS & \textbf{FAIL} & yes \\
Gemini 3.1 Pro & 2 & FAIL & FAIL & PASS & PASS & \textbf{FAIL} & yes \\
Gemini 3.1 Pro & 3 & PASS & PASS & PASS & PASS & \textbf{PASS} & -- \\
Gemini 3.1 Pro & 4 & FAIL & PASS & PASS & FAIL & \textbf{FAIL} & yes \\
\bottomrule
\end{tabular}
\end{table}

The composite PASS rate is 6 of 15. Claude Opus 4.7 passes 3 of 5, GPT-5.5 passes 2 of 5, and Gemini 3.1 Pro passes 1 of 5. The headline number obscures the more interesting structure of the result, which is that the three models fail in three different places.

First, \textbf{each model's failures are concentrated on a different cognitive move}. Claude PASSes Stage 1 in every trial and the structural axis in every trial. Its two FAILs come from Stage 2, where the operational rewrite of the induced rules drops or distorts a rule that was correctly stated at Stage 1. GPT-5.5 PASSes the entire content axis (Stage 1, Stage 2, Stage 3) in 5 of 5 trials but FAILs the structural axis in 3 of 5: the rules say the right thing one by one, yet the rule set as a whole is bloated, redundant, or lacks the cross-rule references the axiomatization instruction asked for. Gemini fails the opposite way: Stage 1 PASSes in only 1 of 5 trials. In each of the other four, the model reimports $F=ma$ from the observations rather than inducing $F=mv$. When Stage 1 does pass (trial 3), the rest of the protocol clears cleanly. The capability gap visible in this framework is therefore not one-dimensional: which of the three moves (induction, formulation, organization) a model fails on differs by model.

Second, across the 45 Stage 3 quantitative-scenario predictions (15 trials, three quantitative scenarios per trial), 0 are direction-correct-but-ratio-leaked. When a model commits to the $F=mv$ direction at Stage 3, it also computes the $F=mv$ ratio rather than the $F=ma$ ratio. The failure mode we worried about most (a model that says the right physics in words and computes the wrong physics in numbers) does not appear in this framework.

\subsection{Takeaway}\label{sec:fmv:summary}

The $F=mv$ result sits in a more nuanced place than a single composite number suggests. Composite PASS is 6 of 15, or 40\%, broken down as Claude 3 of 5, GPT 2 of 5, and Gemini 1 of 5. The more informative split is in \textbf{where} each model fails. Claude is the strongest induction-plus-organization combination but loses 2 of 5 trials at Stage 2 formulation. GPT-5.5 has the right rules in every trial yet writes them as a redundant, unstructured rule set in 3 of 5 trials. Gemini barely induces the framework at all, with four of five trials reimporting $F=ma$ from the observations. Three models, three different cognitive bottlenecks. Which of the three moves (induction, formulation, organization) a model fails on is the per-model finding, not whether it passes overall.

$F=mv$ is the shallowest of the three frameworks in this paper: a single equation, a single domain, observations that already imply the rule. Even here, the composite PASS rate is 40\%, and the failures are not noise but distinct, model-specific cognitive bottlenecks. In the next two frameworks, where conceptual depth, domain count, and historical embedding all rise at once, the same models behave very differently.

\section{Medium: Aristotelian Mechanics}\label{sec:aristotelian}

\subsection{Framework and observations}\label{sec:aristotelian:intro}

Aristotelian mechanics is a real historical framework, not a counterfactual we constructed. Its core claims diverge from Newtonian physics on several axes at once: heavier bodies fall faster in proportion to weight, the medium through which a body moves slows it down further (a stone falls quickly through air and slowly through honey), bodies seek their natural place (heavy substances downward, fire and smoke upward), forced motion ceases when its mover is removed (a cart slows and stops once the push ends), and the celestial regime (the Sun, the Moon, the fixed stars on circular paths) is exempt from the terrestrial rules. Vacuum is rejected as physically impossible~\citep{lloyd1968aristotle}. Two layers of difficulty sit on top of \S\ref{sec:fmv}'s single-equation counterfactual. The framework is multi-principle rather than single-equation, and it is historically real, so training data contains it overwhelmingly in the form of ``Aristotle was wrong''. A model reasoning inside the framework must suspend the rebuttal sentence its training data overwhelmingly endorses.

As with $F=mv$, the framework's name appears nowhere in any prompt the model sees. The model receives 12 hand-written observations describing what an everyday observer would see, with no underlying rule stated and no algebra. A representative observation reads, verbatim from Appendix~\ref{app:aristotelian-details}: \emph{A wooden cart is pushed along a level dirt road. While the pusher's hands remain on the cart it continues to roll. Once the pusher lets go, the cart slows and within a short distance comes to rest.} The banned-word list shifts to match this framework: \emph{inertia}, \emph{acceleration}, \emph{force} (as a defined quantity), \emph{momentum}, \emph{energy}, \emph{mass} (as distinct from weight), \emph{density}, \emph{gravity}, \emph{vacuum}, \emph{friction}, and the proper name of any post-Aristotelian physicist (\emph{Newton}, \emph{Galileo}, \emph{Archimedes}, \ldots). The test is purely lexical and applied to the whole response, just as in \S\ref{sec:fmv:intro}.

The Stage 1 instruction is the same axiomatization-style prompt described in \S\ref{sec:fmv:intro}. The model is asked to propose a self-consistent set of rules that explains every observation, using only the descriptive vocabulary the observations themselves use, returned as a numbered list, in the smallest such set, with any rule that follows from another marked explicitly. As with $F=mv$, the framework rule is what the model has to find on its own. Producing a rule set equivalent to the Aristotelian framework is a Stage 1 PASS. Producing a rule set that reimports Newtonian categories, by name or by content, is a Stage 1 FAIL.

The content axis is eight necessary conditions N1--N8 on the induced rules. N1 requires the two-regime distinction for terrestrial motion: forced motion that ceases when its mover is removed, versus motion or rest that is the body's default state. N2 requires the heavier-falls-faster ordering. N3 requires medium-resistance dependence: a thicker substance slows a moving body more than a thinner one. N4 requires shape dependence at equal weight: a compact ball reaches the ground before a flat sheet of the same weight. N5 requires directional preference of substances: smoke and flame move upward of their own accord, water and stone downward, regardless of how the surrounding body is oriented. N6 requires the heaven/earth split: the Sun, the Moon, and the fixed stars on their unending circular paths cannot be subsumed under the terrestrial rules. N7 requires acknowledgment of the projectile tension: the arrow continuing in flight after leaving the bowstring is in tension with N1 and must either be flagged as unresolved or resolved by an impetus-style account in which the impressed motion explicitly fades~\citep{clagett1959mechanics}. N8 requires some account of floating. Failing any one of N1--N8 is a Stage 1 FAIL. The framework documents a non-exhaustive list of near-pass patterns showing how Newtonian categories typically leak in (\emph{e.g.}\ ``heavier falls faster because of greater gravitational force'', ``the arrow continues because of momentum carried from the bow'', ``flame rises because hot air is less dense''). The full list is in Appendix~\ref{app:aristotelian-details}.

The structural axis N9--N12 is identical to the $F=mv$ structural axis described in \S\ref{sec:fmv:intro}: parsimony, independence, traceability, and hierarchy of the Stage 1 rule set, with thresholds in Appendix~\ref{app:fmv-details}. A trial's composite verdict is again content $\land$ structural: it PASSes only if all three content stages PASS and the structural axis PASSes.

The tested models are the same three: Claude Opus 4.7, GPT-5.5, and Gemini 3.1 Pro.

\subsection{Results}\label{sec:aristotelian:results}

We ran $N=5$ trials per model, for 15 trials in total. Table~\ref{tab:aristotelian-trials} gives the post-audit per-trial classification. The Over-claim column reports the Stage 4 metacognitive axis and is discussed for all three frameworks together in \S\ref{sec:cross-framework:overclaim}.

\begin{table}[h]
\centering
\caption{Post-audit per-trial classification for the Aristotelian framework. $N=5$ trials per model, 15 trials total. Stage 1, Stage 2, Stage 3 are the content-axis stages judged on N1--N8. Structural is the rule-set axis judged on N9--N12 (Appendix~\ref{app:aristotelian-details}). Composite is content $\land$ structural: a trial PASSes only if Stage 1, Stage 2, Stage 3, and Structural all PASS. Over-claim is the separate Stage 4 metacognitive axis (\S\ref{sec:cross-framework:overclaim}): \texttt{yes}/\texttt{no} which is only defined on the content-FAIL trials. \\}
\label{tab:aristotelian-trials}
\small
\begin{tabular}{llcccccc}
\toprule
\textbf{Model} & \textbf{Trial} & \textbf{Stage 1} & \textbf{Stage 2} & \textbf{Stage 3} & \textbf{Structural} & \textbf{Composite} & \textbf{Over-claim} \\
\midrule
Claude Opus 4.7 & 0 & FAIL & FAIL & FAIL & PASS & \textbf{FAIL} & yes \\
Claude Opus 4.7 & 1 & PASS & PASS & FAIL & PASS & \textbf{FAIL} & yes \\
Claude Opus 4.7 & 2 & PASS & FAIL & PASS & PASS & \textbf{FAIL} & no \\
Claude Opus 4.7 & 3 & PASS & PASS & PASS & PASS & \textbf{PASS} & -- \\
Claude Opus 4.7 & 4 & PASS & PASS & PASS & PASS & \textbf{PASS} & -- \\
\midrule
GPT-5.5 & 0 & PASS & PASS & PASS & PASS & \textbf{PASS} & -- \\
GPT-5.5 & 1 & PASS & PASS & PASS & PASS & \textbf{PASS} & -- \\
GPT-5.5 & 2 & PASS & PASS & PASS & PASS & \textbf{PASS} & -- \\
GPT-5.5 & 3 & PASS & PASS & FAIL & PASS & \textbf{FAIL} & yes \\
GPT-5.5 & 4 & PASS & PASS & PASS & PASS & \textbf{PASS} & -- \\
\midrule
Gemini 3.1 Pro & 0 & FAIL & PASS & FAIL & PASS & \textbf{FAIL} & yes \\
Gemini 3.1 Pro & 1 & FAIL & FAIL & FAIL & PASS & \textbf{FAIL} & yes \\
Gemini 3.1 Pro & 2 & FAIL & FAIL & FAIL & PASS & \textbf{FAIL} & yes \\
Gemini 3.1 Pro & 3 & PASS & FAIL & PASS & PASS & \textbf{FAIL} & no \\
Gemini 3.1 Pro & 4 & FAIL & FAIL & PASS & PASS & \textbf{FAIL} & yes \\
\bottomrule
\end{tabular}
\end{table}

The composite PASS rate is 6 of 15. Claude Opus 4.7 passes 2 of 5, GPT-5.5 passes 4 of 5, and Gemini 3.1 Pro passes 0 of 5. The headline number matches \S\ref{sec:fmv} at 6 of 15, but the structure of failure is different. Three observations from the data stand out.

First, \textbf{the structural axis is no longer the bottleneck}. All 15 trials PASS the structural axis. Where in \S\ref{sec:fmv} the axiomatization instruction left structural failures (especially for GPT-5.5, 2 of 5), here it covers the structural axis uniformly. The composite verdict is therefore determined entirely on the content axis, and the model ordering reflects it: GPT 4 of 5, Claude 2 of 5, Gemini 0 of 5. Gemini fails at Stage 1 on 4 of 5 trials, importing Newtonian categories instead of inducing the Aristotelian rule set, and on the fifth trial (trial 3) where Stage 1 does PASS, Stage 2 drops a rule and the trial still fails.

Second, the content-axis failures cluster into four recognisable types of Newton leak, all observable in the audited disagreements: (a) Banned-vocabulary derivatives: Claude trial 0 induced rules referring to bodies as ``denser'', encoding the banned concept of density without using the word itself, and Claude trial 2 introduced ``heavier per equal volume'' at Stage 2 (the operational definition of density, dressed differently). (b) Training-data concept import: Claude trial 0 Stage 2 wrote that a body ``speeds up, slows down, or holds steady'', importing the concept of acceleration that no observation supports (no observation describes a fall-speed changing within a single fall), and Gemini trial 4 Stage 2 said the celestial bodies move at ``constant, unchanging speeds'', a claim Stage 1 had not made. (c) Standard-physics knowledge exposure: Claude trial 0 Stage 3 wrote ``I will not silently import a standard-physics answer (such as that the feather falls just like the iron ball)'', exposing knowledge of the post-Galilean vacuum result while attempting to disclaim it (naming the concept to deny it still counts as use), and Claude trial 1 Stage 3 predicted the feather would fall straight down inside a sealed evacuated chamber, instead of rejecting the vacuum scenario as the framework requires. (d) Missing framework concept: Gemini's Stage 1 never induced antiperistasis or impetus, so its Stage 3 predictions for the arrow defaulted to ``immediately falls strictly straight downward'' (trials 1 and 2). These types show why the lexical banned-token check alone is insufficient: even when the exact banned token is absent, the underlying content can be unmistakably Newtonian, and only the audit catches it.

\subsection{Takeaway}\label{sec:aristotelian:summary}

The Aristotelian result has the same composite PASS rate as $F=mv$, 6 of 15, but the underlying picture is different. The structural axis is uniformly PASSed on this framework (15 of 15), so composite outcomes are determined entirely on the content axis. Per-model split is GPT 4 of 5, Claude 2 of 5, Gemini 0 of 5. Where \S\ref{sec:fmv} showed three different cognitive bottlenecks across the three models, here the bottleneck is the same one across all three: Stage 1 induction in the face of a strong training-data prior. The model ordering reflects ability to suppress that prior. GPT does so most consistently. Claude is in the middle. Gemini cannot, with four of five trials reimporting Newtonian categories at Stage 1 or Stage 2. The content failures themselves cluster into four kinds of Newton leak (\S\ref{sec:aristotelian:results}): banned-vocabulary derivatives, training-data concept imports, standard-physics knowledge exposure, and missing framework concepts. The lexical banned-token check alone does not reliably catch these, but under audit they are unmistakably Newtonian.

Aristotelian sits one level above $F=mv$ on the difficulty gradient. The framework is multi-principle rather than single-equation, and it is historically real, so training data contains it predominantly in the form of a refutation. A model reasoning inside it must suspend a rebuttal sentence its training data overwhelmingly endorses. Even when the structural axis is taken out of the way by the Stage 1 prompt, composite PASS is still 6 of 15, the same headline number as $F=mv$, but the leaks point at a different cognitive limit: ability to resist the trained prior. Decay World (\S\ref{sec:decay}) raises this pressure further by adding multi-domain unification and an absent substrate, and the same models behave very differently there.

\section{Hard: Decay World}\label{sec:decay}

\subsection{Framework and observations}\label{sec:decay:intro}

We also designed Decay World as a counterfactual framework for this paper. It imposes one rule across four physical domains at once: every closed system loses a fixed fraction of its measurable state per second, at the same rate, regardless of domain. A pendulum's amplitude, a sealed cup's temperature, a spinning top's rotation rate, and an orbiting marble's radius all shrink at the same per-second ratio, $0.99$ per second, or $1\%$ loss per second. Gravity, contact, sound, and the rest of ordinary mechanics behave as expected. The only counterfactual is the global slow loss, applied universally. The framework pushes three cognitive pressures at once. First, the rule is counterfactual, as in $F=mv$. Second, the decay has no substrate: it is intrinsic to the system, with no friction, drag, damping, or radiation to attribute it to. Third, the rate is universal across four physical domains, so a mechanical, a thermal, a rotational, and an orbital observation all have to be tied to a single per-second rate.

As in the prior two frameworks, the framework's name and the rule appear nowhere in any prompt the model sees. The model receives 10 hand-written observations describing what a careful observer would directly measure in this world. A representative observation, verbatim from Appendix~\ref{app:decay-details}: \emph{A weight on a spring oscillates back and forth on a perfectly smooth, perfectly level track inside an evacuated chamber. Released with an initial amplitude of 10 cm, the amplitude is measured to be 3.7 cm exactly 100 seconds after release.} The banned-word list is longer than in the previous two frameworks because three layers of vocabulary have to be excluded at once. The energy-and-thermodynamics layer (\emph{energy}, \emph{kinetic}, \emph{potential}, \emph{conservation}, \emph{entropy}, \emph{thermodynamic}, \emph{Hamiltonian}, \emph{Lagrangian}) blocks the model from positing energy as the underlying decaying quantity. The decay-mechanism layer (\emph{friction}, \emph{drag}, \emph{damping}, \emph{dissipation}, \emph{viscous}, \emph{air resistance}, \emph{resistance}, \emph{attenuation}) blocks the model from attributing the slowdown to any contact or medium mechanism. The mechanics layer (\emph{force}, \emph{mass}, \emph{acceleration}, \emph{momentum}, \emph{inertia}) and physicist proper names are also banned. The full list is in Appendix~\ref{app:decay-details}. The same purely-lexical, naming-to-deny-still-counts rule applies.

The Stage 1 instruction is the same axiomatization-style prompt described in \S\ref{sec:fmv:intro}. Unlike $F=mv$ and Aristotelian, where the axiomatization paragraph was added at a structure-prompt round to compare against a no-cue round, Decay World bakes the instruction in from the start. The cue's effect on the structural quality of the rule set was already established by the two prior frameworks. The question for Decay is whether the cued induction can clear a deliberately harder content axis on a multi-domain, no-substrate counterfactual.

The content axis is six necessary conditions N1--N6 on the induced rules. N1 requires closed systems to lose their measurable state over time, on their own, with no external mechanism. N2 requires the decline to be multiplicative (a constant per-second ratio), not additive. N3 requires the rate to be fixed by elapsed time, not by cycle count or by contact. N4 requires the rate to be universal across all closed systems regardless of domain, kind of motion, or measured quantity. N5 requires the rate to be independent of weight, material, and composition. N6 requires the numerical value of the rate to be stated, approximately $0.99$ per second, derivable from the three quantitative observations (spring amplitude, sealed-cup temperature, spinning top rate). A separate list of disqualifying patterns P1--P7 covers contact-mechanism rescue (P1), hidden-substrate framing (P2, the energy-substrate trap), additive decay (P3), per-cycle rate (P4), material-dependent rate (P5), decay without a rate (P6), and refusal of the world (P7). Both N1--N6 and P1--P7 in full are in Appendix~\ref{app:decay-details}.

We do not apply the structural axis (N9--N12, \S\ref{sec:fmv:intro}) to this framework, because the two earlier frameworks already use it to test rule-set organization and Decay is designed to stress a different bottleneck: whether a structurally cued induction can survive multi-domain quantitative application without an underlying substrate. A trial's composite verdict is the content axis alone: PASS at Stage 1, PASS at Stage 2, and PASS at every Stage 3 scenario, where Stage 3 administers four quantitative scenarios and one qualitative scenario per trial. The tested models are the same three: Claude Opus 4.7, GPT-5.5, and Gemini 3.1 Pro.

\subsection{Results}\label{sec:decay:results}

We ran $N=5$ trials per model, for 15 trials in total. Table~\ref{tab:decay-trials} gives the post-audit per-trial classification.
\begin{table}[h]
\centering
\caption{Post-audit per-trial classification for the Decay World framework. $N=5$ trials per model, 15 trials total. Stage 1, Stage 2, Stage 3 are the content-axis stages judged on N1--N6 (Appendix~\ref{app:decay-details}). Stage 3 collapses five per-trial scenarios (four quantitative, one qualitative) by AND. Composite is the content axis alone: a trial PASSes only if all three content columns are PASS. \\}
\label{tab:decay-trials}
\small
\begin{tabular}{llcccccc}
\toprule
\textbf{Model} & \textbf{Trial} & \textbf{Stage 1} & \textbf{Stage 2} & \textbf{Stage 3} & \textbf{Composite} & \textbf{Over-claim} \\
\midrule
Claude Opus 4.7 & 0 & PASS & PASS & FAIL & \textbf{FAIL} & yes \\
Claude Opus 4.7 & 1 & PASS & FAIL & FAIL & \textbf{FAIL} & no \\
Claude Opus 4.7 & 2 & PASS & PASS & FAIL & \textbf{FAIL} & yes \\
Claude Opus 4.7 & 3 & FAIL & PASS & FAIL & \textbf{FAIL} & yes \\
Claude Opus 4.7 & 4 & PASS & FAIL & FAIL & \textbf{FAIL} & no \\
\midrule
GPT-5.5 & 0 & PASS & FAIL & FAIL & \textbf{FAIL} & yes \\
GPT-5.5 & 1 & FAIL & FAIL & FAIL & \textbf{FAIL} & no \\
GPT-5.5 & 2 & PASS & PASS & FAIL & \textbf{FAIL} & yes \\
GPT-5.5 & 3 & FAIL & FAIL & FAIL & \textbf{FAIL} & no \\
GPT-5.5 & 4 & FAIL & FAIL & FAIL & \textbf{FAIL} & yes \\
\midrule
Gemini 3.1 Pro & 0 & PASS & FAIL & FAIL & \textbf{FAIL} & no \\
Gemini 3.1 Pro & 1 & FAIL & FAIL & FAIL & \textbf{FAIL} & yes \\
Gemini 3.1 Pro & 2 & FAIL & FAIL & FAIL & \textbf{FAIL} & yes \\
Gemini 3.1 Pro & 3 & FAIL & FAIL & FAIL & \textbf{FAIL} & yes \\
Gemini 3.1 Pro & 4 & FAIL & FAIL & PASS & \textbf{FAIL} & yes \\
\bottomrule
\end{tabular}
\end{table}

The composite PASS rate is 0 of 15. No trial passes on any model. The headline is not the zero, but the structure of how the zero is reached is more informative. Three observations stand out.

First, \textbf{the failure profile differs by model on the content axis, but every trial fails by Stage 3}. Claude PASSes Stage 1 on 4 of 5 trials, the strongest induction performance of the three models on this framework, then loses 2 of 5 at Stage 2 and every trial at Stage 3. GPT PASSes Stage 1 on only 2 of 5, well below its $F=mv$ and Aristotelian performance, and Stage 2 on 1 of 5. Gemini PASSes Stage 1 on 1 of 5, and Stage 2 on 0 of 5. Of the 8 Stage 1 FAILs across the three models, the most common first-FAIL clause is N4 (universality across domains, 4 of 8 FAILs). Two failures are coverage gaps. One is the hidden-substrate pattern P2, which the pre-registration had predicted would be the most common trap but which occurred only this once. One is N6 (no rate stated). The point that matters: \textit{the rule that the same rate ties four very different domains together} is the hardest condition to induce. Models can describe each domain's decay separately and still fail.

Second, the headline finding is the \textbf{ratio leak on Stage 3 quantitative scenarios}. Stage 3 evaluates four quantitative scenarios plus one qualitative scenario per trial. Across 15 trials that gives 60 quantitative-scenario predictions, and the three-bucket breakdown is:

\begin{itemize}\setlength\itemsep{2pt}
  \item Decay-correct (right direction, right ratio): \textbf{37 of 60} (62\%).
  \item Direction-correct but ratio-leaked (right direction, wrong ratio): \textbf{23 of 60} (38\%).
  \item Direction-wrong: \textbf{0 of 60}.
\end{itemize}

\noindent The direction column is unanimous. No model ever predicts the wrong sign of the change. The 23 ratio-leaked predictions are the failure mode we worried about most across all three frameworks and that did not appear at $F=mv$ (0 of 45 ratio-leaked there) or at Aristotelian (Stage 3 scenarios there are largely qualitative). At Decay it appears 38\% of the time. The typical leak is a model that has correctly named the per-second exponential decay but computes the per-scenario number using a standard-physics relation, most often by treating an unstated energy as the underlying decaying quantity and back-deriving the measured quantity from it (which gives an inconsistent rate across domains). The model's qualitative physics intuition and its quantitative physics computation come apart cleanly here. Across the prior two frameworks, the cosmetic banned-token check passed and the audit caught the leak in words. Here the banned-token check passes and the audit catches the leak in numbers.

Third, the most common Stage 4 over-claim pattern here is the model asserting Stage 1--2 cleared while the trial's Stage 1 or Stage 2 was in fact a FAIL: under self-review, the model defends its induction even when the induction did not actually establish a universal rate. The over-claim rate and the cross-framework comparison are discussed in \S\ref{sec:cross-framework:overclaim}.

\subsection{Takeaway}\label{sec:decay:summary}

The Decay World result reads as 0 of 15 composite, but the more informative story is shape rather than count. The qualitative direction column is perfect: 0 of 60 Stage 3 predictions point the wrong way. The quantitative ratio column is where the framework collapses: 23 of 60 predictions give a ratio derivable only from a standard-physics relation, most often by treating an unstated energy as the underlying decaying quantity, even though no observation supports such a layer and the banned-token list excludes the word. The model knows the world decays. The model cannot compute the decay using the framework's rule. This is the failure mode the prior two frameworks were designed to surface but did not, and it surfaces here cleanly. Stage 1 itself collapses on the universality condition (N4): models can induce that the spring decays, that the cup cools, that the top slows, that the orbit shrinks, without inducing that the four are tied to one rate. The hidden-substrate trap (P2) the prereg flagged as the headline danger fires only once across 8 Stage 1 FAILs. The bigger danger turns out to be one cognitive level below: not what the underlying decaying quantity is, but whether one even exists across domains.

Decay World sits at the floor of the three-framework difficulty gradient. The three cognitive pressures are pushed at the same time: counterfactual reasoning, the absence of a substrate the rule could rest on, and the requirement to unify four physical domains under a single per-time rate. With all three pressures active, the cued axiomatization prompt that lifted composite at $F=mv$ to $6/15$ and at Aristotelian to $6/15$ no longer rescues the content axis here. The next section pulls the methodology-level findings that hold across all three frameworks together: Stage 4 over-claim is consistently high across the three tested frameworks ($67$--$83\%$ of failure-containing trials), LLM judge reliability does not transfer across content / structural / per-scenario tasks within a single framework, and the quantitative ratio leak observed here is the most pointed instance of the qualitative-versus-quantitative asymmetry the three frameworks were jointly designed to probe.

\section{Cross-Framework Findings}\label{sec:cross-framework}

Aligning the experimental results across the three frameworks surfaces a set of findings decoupled from the specific physics content of any one framework. They fall into two kinds. The first kind is empirical regularities that recur across the three frameworks: how often models over-claim under self-review, and which LLM judge is more reliable on which task. The second kind is a methodology contribution: a specific, reproducible LLM-as-judge failure mode surfaced by the Decay World banned-token test, and a retrospective on whether the difficulty-gradient design separated the cognitive moves as intended.

A point of methodology bookkeeping for the rest of this section. The headline composite results reported in Sections~\ref{sec:fmv} and~\ref{sec:aristotelian} come from runs that include the axiomatization paragraph in the Stage 1 prompt described in \S\ref{sec:fmv:intro}. The two cross-framework comparisons in this section are sourced per finding. The Stage 4 over-claim comparison (\S\ref{sec:cross-framework:overclaim}) comes from the same structure-prompt runs as the headline composite, so it is measured on the identical condition across all three frameworks. The judge-reliability comparison (\S\ref{sec:cross-framework:judge}) comes from each framework's first fully audited round, the first instance of the full pre-registration plus dual-judge plus human-audit pathway. Section~\ref{sec:decay} (Decay World) bakes the axiomatization paragraph in from the start, so a single set of runs supports the headline and both comparisons. Appendix~\ref{app:prereg-tags} lists the locked pre-registration tag for each round.

\subsection{Difficulty Gradient in Retrospect}\label{sec:cross-framework:gradient}

\begin{figure}[h]
\centering
\includegraphics[width=0.85\textwidth]{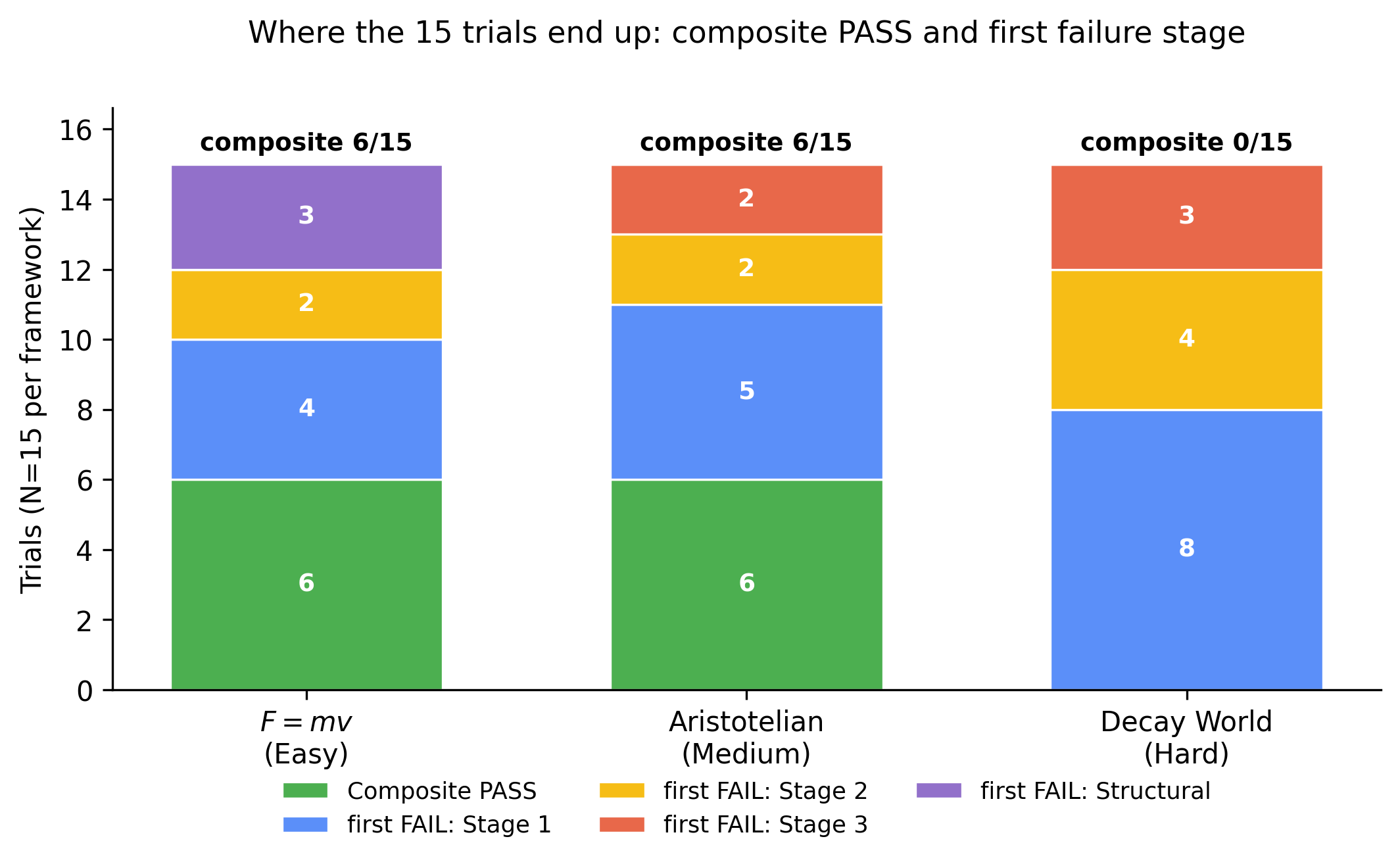}
\caption{Where the 15 trials end up in each framework: composite PASS, or the stage at which the trial first failed (Stage 1 induction, Stage 2 formulation, Stage 3 prediction, or the structural axis). Composite PASS falls across the difficulty gradient at 6/15, 6/15, 0/15. Structural-only failures appear only at $F=mv$. Aristotelian applies the same structural axis and passes it uniformly, while Decay does not apply the structural axis at all. The Decay World bar shows the first failing stage per trial. Its quantitative Stage 3 bottleneck is a separate cumulative pattern, reported in \S\ref{sec:decay} as 14 of 15 trials failing the Stage 3 check and 23 of 60 ratio-leaked quantitative predictions.}
\label{fig:bottleneck}
\end{figure}

The three frameworks were ordered by intended difficulty before any experiment ran (\S\ref{sec:methods:worlds}). The post-audit composite PASS rates 6/15, 6/15, 0/15 do not by themselves confirm a strict monotone difficulty ordering, since the two easier frameworks tie at 6/15 and a reader looking only at the composite count would conclude that $F=mv$ and Aristotelian are equally difficult. The per-stage breakdown indicates they are not equally difficult in the same sense. \textbf{The empirical signature of the gradient is not in the composite count but in where the cognitive bottleneck sits across the three frameworks.} Each framework's failures concentrate in a different cognitive move (Figure~\ref{fig:bottleneck}).

\paragraph{$F=mv$ (Easy).} Failures spread across all three judged moves, with each model failing on a different one. Claude loses 2 of 5 trials at Stage 2 formulation. GPT loses 3 of 5 at the structural axis, induction-clean but with a disorganized rule set. Gemini fails Stage 1 induction in 4 of 5 trials by reimporting $F=ma$. The framework is shallow enough that the per-model capability gap is visible as three different bottlenecks at once.

\paragraph{Aristotelian (Medium).} The dominant bottleneck is induction under a strong training-data prior, surfacing as Stage 1 failures and their Stage 2 inheritance, most visibly in Gemini's four Stage 1 FAILs. The model ordering on composite is driven by each model's ability to suppress the ``Aristotle was wrong'' rebuttal sentence that training data overwhelmingly endorses. The structural axis, which was a major differentiator at $F=mv$, is uniformly PASSed under the axiomatization prompt and contributes nothing to the variance.

\paragraph{Decay (Hard).} The bottleneck shifts again, this time to a different stage. Every Decay trial fails by Stage 3, and 14 of 15 fail the Stage 3 check itself. The qualitative direction column is unanimous across 60 quantitative predictions (0 direction-wrong), but the ratio column is wrong on 23 of 60. The Stage 1 failures that do occur (8 of 15) concentrate on N4 (universality across domains), not on the hidden-substrate trap P2 the prereg flagged as the modal danger.

\begin{table}[h]
\centering
\caption{Cross-framework comparison of composite PASS and primary failure mode. Composite count alone (6/15, 6/15, 0/15) understates the gradient: each framework's failures concentrate in a different cognitive move, and the move shifts as difficulty rises.}
\label{tab:cross-framework-gradient}
\small
\begin{tabular}{llll}
\toprule
\textbf{Framework} & \textbf{Composite PASS} & \textbf{Primary failure stage} & \textbf{Failure mode} \\
\midrule
$F=mv$ & 6/15 & Different per model & Per-model bottleneck heterogeneity \\
Aristotelian & 6/15 & Stage 1 induction & Training-prior leak \\
Decay World & 0/15 & Stage 3 (FAIL in 14 of 15) & Quantitative ratio leak \\
\bottomrule
\end{tabular}
\end{table}

The pattern across the three frameworks reads as a \textbf{bottleneck migration}: from per-model heterogeneity at $F=mv$, to a shared Stage 1 training-prior bottleneck at Aristotelian, to a Stage 3 quantitative-computation bottleneck at Decay. A composite count alone would not show this migration. The decomposition into Stage 1 / Stage 2 / Stage 3 / structural (\S\ref{sec:methods:protocol}) is the instrument that surfaces it.

The axiomatization Stage 1 prompt added at the structure-prompt rounds of $F=mv$ and Aristotelian lifted composite PASS at both frameworks and reduced structural-axis failures at both. At Decay the same prompt was baked in from the start, and it did not rescue the content axis: composite stayed at 0/15. Prompt engineering of this kind helped when the bottleneck was rule-set organization or Stage 1 induction itself, and did not rescue the content axis in our Decay runs, where the bottleneck is quantitative Stage 3 computation. The implication for benchmark design is narrow but pointed: this suggests that a Stage 1 axiomatization cue is insufficient when the bottleneck is quantitative cross-domain computation, however well the same cue performs on easier frameworks.

The retrospective verdict on the design is that the three frameworks did separate the cognitive moves the four-stage protocol was meant to test. $F=mv$ mostly exercised induction and rule-set organization. Aristotelian put nearly all the pressure on induction under training-prior. Decay shifted the pressure to quantitative computation. The composite count alone would have hidden this. The per-stage and per-axis breakdown made it readable.

\subsection{Stage 4 Over-Claim Is Consistently High Across Frameworks}\label{sec:cross-framework:overclaim}

Across the three frameworks, the Stage 4 over-claim rate on failure-containing trials is high in every case and never falls below two-thirds. $F=mv$ records 5 of 6 (83\%), Aristotelian 7 of 9 (78\%), and Decay 10 of 15 (67\%) (Table~\ref{tab:overclaim}). The rates span 67--83\% on widely different denominators (6, 9, 15), all measured on the same structure-prompt arm as the headline composite.

\begin{table}[h]
\centering
\caption{Stage 4 over-claim across the three frameworks, all on the structure-prompt arm. A failure-containing trial is one in which at least one of Stages 1--3 was a FAIL. Over-claim is the subset of those trials whose Stage 4 self-assessment claims that no earlier error occurred. The denominators differ widely (6, 9, 15), yet the rate is at or above two-thirds in every framework.}
\label{tab:overclaim}
\small
\begin{tabular}{lccc}
\toprule
\textbf{Framework} & \textbf{Failure-containing trials} & \textbf{Over-claim trials} & \textbf{Over-claim rate} \\
\midrule
$F=mv$ (Easy) & 6 & 5 & 83\% \\
Aristotelian (Medium) & 9 & 7 & 78\% \\
Decay World (Hard) & 15 & 10 & 67\% \\
\bottomrule
\end{tabular}
\end{table}

What varies across these three frameworks is extensive. Framework difficulty varies from Easy to Hard, with composite PASS rates dropping from 6/15 to 0/15. Framework nature varies between counterfactual ($F=mv$, Decay) and historically real (Aristotelian). The failure modes that the Stage 4 review has to identify vary by framework: training-prior leak at Aristotelian, formulation slips and structural disorganization at $F=mv$, and quantitative ratio leaks at Decay. None of this brings the over-claim rate below two-thirds.

The implication is that Stage 4 self-review is weak in every tested framework, not a framework-specific lapse. Frontier models miss their own just-committed slips in at least two-thirds of failure-containing trials regardless of whether the slip is a Stage 1 induction error against a strong training prior or a Stage 3 quantitative computation in a novel domain. If anything the rate is highest on the easiest framework ($F=mv$, 83\%) and lowest on the hardest (Decay, 67\%), the opposite of what a ``harder tasks surface more difficulty signals'' calibration story predicts, though with $N=6, 9, 15$ we read this as the absence of any reliable difficulty trend rather than a reverse gradient. This is the inverse of the common ``models know what they don't know'' calibration framing, at least within the three frameworks tested here.

This pattern, a model-level meta-cognitive pattern that does not appear to respond to framework difficulty in this dataset, is one component of the integrated portrait \S\ref{sec:discussion:profile} develops from the three experiments combined.

\subsection{Judge Reliability Does Not Transfer Across Frameworks}\label{sec:cross-framework:judge}

Each of the three frameworks triggered the prereg's 25\% disagreement-rate threshold and went through human audit (\S\ref{sec:methods:judge}). The disagreement cases give a controlled head-to-head: on every case where the two LLM judges disagreed, the human audit produced a canonical verdict, and we can measure how often each judge matched it. Table~\ref{tab:cross-framework-judge} reports this comparison for the content-axis disagreement cases of each framework's first audited round.

\begin{table}[h]
\centering
\caption{LLM judge agreement with the human audit on the dual-judge content-axis disagreement cases for each framework's first audited round. The ``more reliable'' judge (bolded per row) reverses between Aristotelian and $F=mv$ and shifts again at Decay.}
\label{tab:cross-framework-judge}
\small
\begin{tabular}{llll}
\toprule
\textbf{Framework} & \textbf{Disagree cases} & \textbf{Claude judge} & \textbf{OpenAI judge} \\
\midrule
Aristotelian & 17 & 3/17 (18\%) & 14/17 (\textbf{82\%}) \\
$F=mv$ & 12 & 10/12 (\textbf{83\%}) & 2/12 (17\%) \\
Decay World & 18 & 12/18 (\textbf{67\%}) & 6/18 (33\%) \\
\bottomrule
\end{tabular}
\end{table}

The pattern is a reversal. On Aristotelian's content axis the OpenAI judge agreed with the human audit on 14 of 17 disagree cases (82\%), and Claude on only 3 of 17. The very next framework, $F=mv$, reverses this completely: Claude matched the audit on 10 of 12 (83\%), and OpenAI on only 2 of 12. Decay shifts the picture again, with Claude leading but at a smaller margin (12 of 18, or 67\%). The two judges are the same models across the three rounds with the same dual-judge architecture. The criteria and prompts differ per framework (each framework has its own Stage 1 prompt and banned-word list), but the architecture is unchanged.

The implication is that the ``more reliable'' LLM judge is framework-dependent and cannot be predicted from prior rounds. A reader who saw only the $F=mv$ judging results would conclude that Claude is the better content judge. A reader who saw only Aristotelian would conclude the opposite. Both readings would be wrong on the next framework. Single-judge LLM evaluation in this paradigm is unsafe in a specific way: not ``judges are sometimes wrong'', which is true and uninteresting, but ``no judge is reliably more reliable across frameworks''. This is direct support for the dual-judge plus disagreement-rate plus audit architecture (\S\ref{sec:methods:judge}). The two judges are needed not as one good and one redundant but as two whose disagreements need a tie-breaker we can trust.

\subsection{A Reproducible LLM-Judge Failure Mode}\label{sec:cross-framework:judge-failure}

The Decay World banned-token test on Stage 1 responses triggered a specific failure in the OpenAI judge that did not appear at comparable magnitude in the prior two frameworks. Asked to check whether a banned token appears in a response, the judge slid from literal substring matching into semantic association. It did this in two ways. The first is fabricated citation: the judge reports a banned token that appears nowhere in the response, having associated the rule's content with the banned concept and reified the association as a literal quote. The second is misclassification: the judge flags a word that is genuinely in the response but is not on the ban list, broadening the explicit lexical test into an implicit semantic one. On audit, 16 of the 18 content-axis disagreement verdicts from the OpenAI judge were defective in one of these two ways.

Three conditions distinguish Decay from the frameworks where the failure stayed rare. First, the banned-word list is long, 20+ tokens across three layers, against 8 at $F=mv$ and 11 at Aristotelian. Second, the topical overlap is high: the observations describe the absence of decay, damping, and dissipation, so almost any rule content sits one semantic step from a banned token. Third, the instruction asks for verbatim token matching, and the slide is precisely from that explicit lexical task to an implicit semantic one. The long list and the topical overlap together make every semantic association look like a plausible literal hit.

The two forms are not equally catchable. A post-hoc check that requires every cited banned token to appear verbatim in the response catches fabricated citations automatically, but it cannot catch misclassification, where the cited word is real. At Decay the misclassifications were the majority, and they were caught only by the dual-judge plus disagreement-rate plus human-audit pathway (\S\ref{sec:methods:judge}), which surfaces both forms as judge-judge disagreement and resolves them against the audit canonical. This is a specific, reproducible failure, not generic judge noise. Its triggers are known, the fabricated form has an automated catch, and the audit architecture catches both. The mechanism is not specific to OpenAI, so we cannot assume Claude would resist a comparably heavy, topically overlapping ban list, even though it did not show the failure at Decay.

\section{Discussion}\label{sec:discussion}

\subsection{Portrait of Model Physics Literacy from the Three Experiments Combined}\label{sec:discussion:profile}

The three experiments, taken together, sketch a portrait of frontier LLM physics literacy along the four cognitive moves (induction, formulation, prediction, review) that the four-stage protocol of \S\ref{sec:methods:protocol} was designed to isolate.

\textbf{Qualitative direction is strong, quantitative computation breaks under cross-domain pressure.} Across the 45 quantitative Stage 3 predictions at $F=mv$ and the 60 at Decay, models almost never predict the wrong sign of the change. $F=mv$ has 0 of 45 direction-wrong predictions. Decay has 0 of 60. In the two frameworks with explicit quantitative Stage 3 counts, the qualitative direction column is essentially solved. The ratio column is not. $F=mv$ has 0 of 45 direction-correct-but-ratio-leaked predictions. Decay has 23 of 60. The asymmetry surfaces only when a single rule has to be applied across four physical domains: the model knows the world decays and gets the sign right, but it cannot do the per-domain quantitative computation under the framework's own rule. The typical failure path is to back-derive the measured quantity from an unstated ``energy'' substrate that the banned-token list excludes. The model's qualitative physical intuition and its quantitative physical computation come apart under cross-domain pressure.

\textbf{Induction is framework-specific, not a stable model property.} Stage 1 pass rates per model swing widely across the three frameworks. At $F=mv$, Claude and GPT both pass Stage 1 cleanly (5 of 5), and the bottleneck for Gemini is reimporting $F=ma$. At Aristotelian, the bottleneck is the strength of the training-data prior. The model ordering on the composite is driven by the ability to suppress the standard-physics corrective prior that treats Aristotelian mechanics as something to repair. At Decay, the bottleneck shifts again to N4 (universality across domains): models can describe each domain's decay separately and still fail to induce that the four are tied to one rate. Three frameworks, three different Stage 1 failure causes. ``Can frontier LLMs induce a physics rule from observations'' is not a question with a single answer. The answer depends on which prior the framework is asking the model to suspend.

\textbf{Rule-set organization is prompt-responsive, not a default capability.} The structural axis (parsimony, independence, traceability, hierarchy) was a major differentiator at $F=mv$. GPT passed every content stage cleanly but failed the structural axis on every trial under the original Stage 1 prompt. After adding one paragraph asking for the smallest rule set with explicit cross-references, the structural pass rate doubled at $F=mv$ and saturated at 15 of 15 at Aristotelian. The capability to organize a rule set into an axiomatic structure is in the model, but the model does not exhibit it by default. The implication is narrow but pointed: a benchmark that judges rule-set quality without an axiomatization cue is measuring not ``can the model organize'' but ``does the model organize without being asked'', and the answer to the second question is largely no.
  
\textbf{Meta-cognitive review is weak across all three tested frameworks, not difficulty-responsive.} Stage 4 over-claim is at or above two-thirds in every framework (67--83\%, \S\ref{sec:cross-framework:overclaim}), with no reliable improvement on harder frameworks. Frontier models miss what they have just done wrong at a consistently high rate, regardless of whether the wrong move was a Stage 1 induction error against a strong training prior or a Stage 3 quantitative computation in a novel domain. The common ``models know what they don't know'' calibration framing predicts the opposite: harder tasks should produce more obvious internal signals of difficulty, which should improve self-assessment. We do not see that. The most charitable reading is that Stage 4 self-review accesses the same internal state as Stages 1--3, so a model that did not notice its slip at Stage 2 cannot notice it at Stage 4 either. The less charitable reading is that an over-claim rate at or above two-thirds in every framework is consistent with a model-level limit on Stage 4 self-review in this paradigm. The data support either reading and argue against the calibration framing for these three frameworks. They do not by themselves distinguish ``ceiling'' from ``coincidence across three data points''.

\subsection{Limitations}\label{sec:discussion:limitations}

The findings above rest on the experimental scope detailed in \S\ref{sec:methods} and Sections~\ref{sec:fmv}--\ref{sec:decay}. Six limitations bound what we can claim.

\begin{itemize}
  \item \textbf{$N=5$ trials per model per framework is descriptive, not statistical inference.} The per-model composite counts are too thin to support ``model X is reliably better than model Y at framework Z'' at any specified confidence level. Our claims are about patterns observed in the 45 trials we ran, not about a population of trials we did not run.
  \item \textbf{Three frameworks do not span the difficulty gradient.} We have an Easy, a Medium, and a Hard. We do not have multiple frameworks at any single difficulty level, so the ``bottleneck migrates with difficulty'' claim of \S\ref{sec:cross-framework:gradient} is consistent with three data points but not directly tested against an alternative ordering of bottlenecks within the same difficulty.
  \item \textbf{Three frontier models are not exhaustive.} Our claims apply to Claude Opus 4.7, GPT-5.5, and Gemini 3.1 Pro at the version strings pinned per round. A reasoning-trained or physics-fine-tuned model could behave differently. We have not tested any open-weight model. The Stage 4 over-claim pattern is across three frontier models in a single time window.
  \item \textbf{Dual judges plus a disagreement-rate threshold plus human audit is an auditable baseline, not a maximally strict one.} A stricter methodology could use three or more judges, sharper criteria, or per-stage human audit. We use this baseline because it is reproducible and verifiable, and a reader who wants a tighter ceiling can re-run from the prereg tags with additional judges. The two judges, Claude and GPT, also share model families with two of the three tested models, so each can score outputs from its own family. The cross-provider judge pair and the human-audit backstop limit a self-preference effect that we do not separately quantify.
  \item \textbf{The axiomatization prompt is one cue among many.} Different prompts (chain-of-thought, debate, decomposition) might surface different bottlenecks or rescue Stage 3 quantitative computation at Decay. Our claim that prompt engineering reaches a ceiling at quantitative cross-domain computation is specific to this cue.
  \item \textbf{Stage 4 over-claim rests on three data points.} Three rates all at or above two-thirds is suggestive of a model-level pattern, but a fourth framework where over-claim drops well below two-thirds is the falsification test, and we have not run it.
\end{itemize}

\subsection{Implications for Benchmark Design}\label{sec:discussion:benchmark}

The three experiments and the cross-framework findings (\S\ref{sec:cross-framework}) carry several implications for how LLM evaluation benchmarks, especially in physics and physics-adjacent reasoning, should be designed.

A per-question or composite PASS rate is insufficient for cognitive evaluation on its own. Composite PASS was 6 of 15 at both $F=mv$ and Aristotelian, yet the cognitive pictures behind those identical counts are completely different: per-model heterogeneity at $F=mv$, a shared training-prior bottleneck at Aristotelian. A benchmark that reports only aggregate accuracy hides exactly the structural differences in capability and failure that are the actual finding. The remedy is to decompose the cognitive chain rather than score it end to end. The four-stage protocol's per-stage verdicts isolate which move failed on which trial, where a benchmark that asks for a single final answer cannot separate an induction failure from an operational-formulation failure from a prediction failure, and the same total accuracy can hide three different cognitive boundaries.

The judging infrastructure is part of the experimental design, not a post-hoc implementation detail. Pre-registration is a floor and not a ceiling: SHA-256-sealed criteria and locked prompts close off one class of post-hoc rationalization, but they do nothing about LLM-judge disagreement or judge fabrication, so an audit pathway triggered by an explicit disagreement threshold becomes necessary as soon as the judging runs at a scale no longer reviewable by hand. The more reliable LLM judge reverses between frameworks (\S\ref{sec:cross-framework:judge}), so a single-judge benchmark would have produced systematically wrong content verdicts on at least one of our three rounds, and the reproducible fabrication failure documented in \S\ref{sec:cross-framework:judge-failure} is one that a single-judge design has no internal mechanism for detecting.

The most important benchmark-design choice is the choice of frameworks themselves. Counterfactual frameworks separate reasoning from recitation, consistent with prior work on counterfactual task variants~\citep{wu2024reasoning}, and for physics specifically a multi-domain counterfactual with no underlying substrate, Decay World here, probes a capability that single-equation counterfactuals like $F=mv$ and historical frameworks like Aristotelian do not reach. The most informative feature was not pass-rate ordering alone, but bottleneck migration across the gradient: a benchmark whose frameworks sit at similar composite PASS but whose failure stage migrates between them carries more information than one whose pass rates fall linearly while the failure stage stays put. Our three frameworks demonstrate this more by accident than by design, and an intentional version of the principle would tighten the next round of physics evaluations.

The takeaway for benchmark design, restated in one sentence: a physics-reasoning benchmark earns its name when it can tell the reader which of the four cognitive moves the model failed on, not when it can report another decimal of accuracy. The four-stage protocol described in this paper is one instance of that principle.

\section{Conclusion}\label{sec:conclusion}

This paper tested three frontier large language models (Claude Opus 4.7, GPT-5.5, Gemini 3.1 Pro) on three parallel physics worlds of increasing difficulty: $F=mv$ (Easy, a single-equation counterfactual), Aristotelian mechanics (Medium, a historical framework), and Decay World (Hard, a four-domain counterfactual). The test asked whether these models can complete the four cognitive moves of the hypothetico-deductive tradition (induction, formulation, prediction, review) inside a physics framework whose conclusions conflict with their training prior. Each framework's pre-registration was SHA-256-sealed, the four-stage protocol enforced fresh API sessions between stages with no context reuse, dual LLM judges produced independent verdicts at each stage, and human audit served as the canonical tie-breaker whenever inter-judge disagreement exceeded a pre-registered threshold. The post-audit composite PASS rates were 6 of 15, 6 of 15, and 0 of 15 across the three frameworks (content $\land$ structural for $F=mv$ and Aristotelian, content axis only for Decay World where the structural axis is out of scope).

The most pointed finding is the asymmetry between qualitative direction and quantitative computation. Across 60 quantitative Stage 3 predictions at Decay World, no model ever predicted the wrong sign of the change, but 23 of the 60 predictions gave a ratio derivable only from a standard-physics relation rather than from the framework's stated rule. The same asymmetry was absent at $F=mv$ (0 of 45 ratio-leaked). This is the failure mode the three frameworks were jointly designed to surface and that only the multi-domain framework actually elicits. Three methodology contributions accompany this finding. First, decomposing the cognitive chain into four independent stages reveals failure structure that the composite count alone hides: $F=mv$ and Aristotelian both score 6 of 15, but the failures concentrate on three different cognitive moves at $F=mv$ and on a shared training-prior bottleneck at Aristotelian. Second, the Stage 4 over-claim rate is at or above two-thirds across all three frameworks (67--83\%), while the ``more reliable'' LLM judge against the human audit reverses between Aristotelian and $F=mv$. The first regularity is suggestive of a frontier-model pattern within this dataset, the second a methodological observation about LLM-as-judge, and both are visible only across more than one framework. Third, the Decay World banned-token test surfaced a specific reproducible failure mode in one of the two judges, whose fabricated-citation form is mechanically checkable and whose broader pattern is guarded by the dual-judge plus audit pathway.

The design principle that emerges, restated in different terms from \S\ref{sec:discussion:benchmark}: a physics-reasoning benchmark should aim to tell its reader which of the four cognitive moves the model failed on, not to report another decimal of aggregate accuracy. A benchmark in which two frameworks share a composite score but place the failure on different cognitive moves carries more cognitive-evaluation information than three accuracies dropping linearly while the failure stage stays constant. For $F=mv$ and Aristotelian, the headline capability results in this paper come from the structure-prompt arms (runs with the axiomatization paragraph in the Stage 1 prompt), while the judge-reliability comparison in \S\ref{sec:cross-framework} comes from each framework's first audited round and the Stage 4 over-claim comparison comes from the same structure-prompt runs as the composite. Decay World supports all of these from a single arm. These claims rest on three frameworks, three frontier models, and $N=5$ trials per cell, and they are descriptive of patterns in the 45 trials we ran rather than statistical inferences over a larger population. Whether the cross-domain quantitative-computation bottleneck at Decay can be lifted by reasoning-trained or physics-fine-tuned models, and whether Stage 4 over-claim stays above two-thirds at a fourth framework outside the present scope, are open questions for the next round of work.

\section*{Code and Data Availability}\label{sec:availability}

All pre-registration files, raw prompts, model responses, dual-judge verdicts, human-audit records, framework artifacts, and reproducibility scripts are at the project repository: \url{https://github.com/dongzhang84/physlit}. The canonical reproducibility recipe is maintained at the head of \texttt{README.md} under the section ``Reproducing the experiments'', kept in sync with the latest pre-registration tags. The three headline pre-registration tags listed in Appendix~\ref{app:prereg-tags} (\texttt{prereg-02\_fmv.2-locked}, \texttt{prereg-v0.3-locked}, \texttt{prereg-03\_decay-locked}) are the canonical sources for the headline composite PASS rates reported in Sections~\ref{sec:fmv}, \ref{sec:aristotelian}, and \ref{sec:decay}. Tested-model output is non-deterministic across vendors under the provider-specific sampling controls described in Section~\ref{sec:methods:config}, so trial responses are not byte-identical to ours, but the post-audit verdict patterns are robust under the audit-replay scripts committed to the repository.

\bibliographystyle{tmlr}
\bibliography{references}

\begin{thebibliography}{48}
\providecommand{\natexlab}[1]{#1}
\providecommand{\url}[1]{\texttt{#1}}
\expandafter\ifx\csname urlstyle\endcsname\relax
  \providecommand{\doi}[1]{doi: #1}\else
  \providecommand{\doi}{doi: \begingroup \urlstyle{rm}\Url}\fi

\bibitem[Alon et~al.(2026)Alon, Bloom, Gowers, Litt, Sawin, Shankar, Tsimerman,
  Wang, and Wood]{openai2026erdos}
Noga Alon, Thomas~F. Bloom, W.~T. Gowers, Daniel Litt, Will Sawin, Arul
  Shankar, Jacob Tsimerman, Victor Wang, and Melanie~Matchett Wood.
\newblock Remarks on the disproof of the unit distance conjecture.
\newblock \emph{arXiv preprint arXiv:2605.20695}, 2026.
\newblock Human-verified companion to the {OpenAI}-generated counterexample.

\bibitem[Battaglia et~al.(2013)Battaglia, Hamrick, and
  Tenenbaum]{battaglia2013simulation}
Peter~W. Battaglia, Jessica~B. Hamrick, and Joshua~B. Tenenbaum.
\newblock Simulation as an engine of physical scene understanding.
\newblock \emph{Proceedings of the National Academy of Sciences (PNAS)},
  110\penalty0 (45):\penalty0 18327--18332, 2013.
\newblock \doi{10.1073/pnas.1306572110}.

\bibitem[Bear et~al.(2021)Bear, Wang, Mrowca, Binder, Tung, Pramod, Holdaway,
  Tao, Smith, Sun, Fei-Fei, Kanwisher, Tenenbaum, Yamins, and
  Fan]{bear2021physion}
Daniel~M. Bear, Elias Wang, Damian Mrowca, Felix~J. Binder, Hsiau-Yu~Fish Tung,
  R.~T. Pramod, Cameron Holdaway, Sirui Tao, Kevin Smith, Fan-Yun Sun,
  Li~Fei-Fei, Nancy Kanwisher, Joshua~B. Tenenbaum, Daniel L.~K. Yamins, and
  Judith~E. Fan.
\newblock Physion: Evaluating physical prediction from vision in humans and
  machines.
\newblock In \emph{NeurIPS Datasets and Benchmarks}, 2021.

\bibitem[Boiko et~al.(2023)Boiko, MacKnight, Kline, and
  Gomes]{boiko2023autonomous}
Daniil~A. Boiko, Robert MacKnight, Ben Kline, and Gabe Gomes.
\newblock Autonomous chemical research with large language models.
\newblock \emph{Nature}, 624:\penalty0 570--578, 2023.
\newblock \doi{10.1038/s41586-023-06792-0}.

\bibitem[Brooks et~al.(2024)Brooks, Peebles, Holmes, DePue, Guo, Jing, Schnurr,
  Taylor, Luhman, Luhman, Ng, Wang, and Ramesh]{brooks2024sora}
Tim Brooks, Bill Peebles, Connor Holmes, Will DePue, Yufei Guo, Li~Jing, David
  Schnurr, Joe Taylor, Troy Luhman, Eric Luhman, Clarence Ng, Ricky Wang, and
  Aditya Ramesh.
\newblock Video generation models as world simulators.
\newblock OpenAI technical report,
  \url{https://openai.com/research/video-generation-models-as-world-simulators},
  2024.

\bibitem[Brown et~al.(2020)Brown, Mann, Ryder, Subbiah, Kaplan, Dhariwal,
  Neelakantan, Shyam, Sastry, Askell, Agarwal, Herbert-Voss, Krueger, Henighan,
  Child, Ramesh, Ziegler, Wu, Winter, Hesse, Chen, Sigler, Litwin, Gray, Chess,
  Clark, Berner, McCandlish, Radford, Sutskever, and Amodei]{brown2020language}
Tom~B. Brown, Benjamin Mann, Nick Ryder, Melanie Subbiah, Jared Kaplan,
  Prafulla Dhariwal, Arvind Neelakantan, Pranav Shyam, Girish Sastry, Amanda
  Askell, Sandhini Agarwal, Ariel Herbert-Voss, Gretchen Krueger, Tom Henighan,
  Rewon Child, Aditya Ramesh, Daniel~M. Ziegler, Jeffrey Wu, Clemens Winter,
  Christopher Hesse, Mark Chen, Eric Sigler, Mateusz Litwin, Scott Gray,
  Benjamin Chess, Jack Clark, Christopher Berner, Sam McCandlish, Alec Radford,
  Ilya Sutskever, and Dario Amodei.
\newblock Language models are few-shot learners.
\newblock In \emph{Advances in Neural Information Processing Systems
  (NeurIPS)}, 2020.

\bibitem[Carlini et~al.(2023)Carlini, Ippolito, Jagielski, Lee, Tram{\`e}r, and
  Zhang]{carlini2023quantifying}
Nicholas Carlini, Daphne Ippolito, Matthew Jagielski, Katherine Lee, Florian
  Tram{\`e}r, and Chiyuan Zhang.
\newblock Quantifying memorization across neural language models.
\newblock In \emph{ICLR}, 2023.
\newblock arXiv:2202.07646.

\bibitem[Clagett(1959)]{clagett1959mechanics}
Marshall Clagett.
\newblock \emph{The Science of Mechanics in the Middle Ages}.
\newblock University of Wisconsin Press, Madison, 1959.

\bibitem[Cobbe et~al.(2021)Cobbe, Kosaraju, Bavarian, Chen, Jun, Kaiser,
  Plappert, Tworek, Hilton, Nakano, Hesse, and Schulman]{cobbe2021training}
Karl Cobbe, Vineet Kosaraju, Mohammad Bavarian, Mark Chen, Heewoo Jun, Lukasz
  Kaiser, Matthias Plappert, Jerry Tworek, Jacob Hilton, Reiichiro Nakano,
  Christopher Hesse, and John Schulman.
\newblock Training verifiers to solve math word problems.
\newblock \emph{arXiv preprint arXiv:2110.14168}, 2021.

\bibitem[Einstein(1916)]{einstein1916grundlage}
Albert Einstein.
\newblock Die {Grundlage} der allgemeinen {Relativit\"atstheorie}.
\newblock \emph{Annalen der Physik}, 49:\penalty0 769--822, 1916.
\newblock General Relativity field equations; predicts light bending in a
  gravitational field, confirmed by the 1919 eclipse.

\bibitem[Ghareeb et~al.(2025)Ghareeb, Chang, Mitchener, Yiu, Szostkiewicz,
  Laurent, Razzak, White, Hinks, and Rodriques]{ghareeb2026robin}
Ali~Essam Ghareeb, Benjamin Chang, Ludovico Mitchener, Angela Yiu, Caralyn~J.
  Szostkiewicz, Jon~M. Laurent, Muhammed~T. Razzak, Andrew~D. White,
  Michaela~M. Hinks, and Samuel~G. Rodriques.
\newblock Robin: A multi-agent system for automating scientific discovery.
\newblock \emph{arXiv preprint arXiv:2505.13400}, 2025.

\bibitem[Gottweis et~al.(2025)Gottweis, Weng, Daryin, Tu, Palepu,
  et~al.]{gottweis2026coscientist}
Juraj Gottweis, Wei-Hung Weng, Alexander Daryin, Tao Tu, Anil Palepu, et~al.
\newblock Towards an {AI} co-scientist.
\newblock \emph{arXiv preprint arXiv:2502.18864}, 2025.

\bibitem[Grattafiori et~al.(2024)]{grattafiori2024llama3}
Aaron Grattafiori et~al.
\newblock The {Llama 3} herd of models.
\newblock \emph{arXiv preprint arXiv:2407.21783}, 2024.

\bibitem[Ha \& Schmidhuber(2018)Ha and Schmidhuber]{ha2018world}
David Ha and J\"{u}rgen Schmidhuber.
\newblock World models.
\newblock In \emph{Advances in Neural Information Processing Systems
  (NeurIPS)}, 2018.

\bibitem[Hassabis(2026)]{hassabis2026einstein}
Demis Hassabis.
\newblock Remarks on an ``{Einstein Test}'' for {AGI}.
\newblock India AI Impact Summit, February 17, 2026, 2026.
\newblock Public address.

\bibitem[Hempel(1966)]{hempel1966philosophy}
Carl~G. Hempel.
\newblock \emph{Philosophy of Natural Science}.
\newblock Prentice-Hall, Englewood Cliffs, NJ, 1966.

\bibitem[Hendrycks et~al.(2021)Hendrycks, Burns, Kadavath, Arora, Basart, Tang,
  Song, and Steinhardt]{hendrycks2021measuring}
Dan Hendrycks, Collin Burns, Saurav Kadavath, Akul Arora, Steven Basart, Eric
  Tang, Dawn Song, and Jacob Steinhardt.
\newblock Measuring mathematical problem solving with the {MATH} dataset.
\newblock In \emph{NeurIPS Datasets and Benchmarks}, 2021.

\bibitem[Huang \& Chang(2023)Huang and Chang]{huang2023reasoning}
Jie Huang and Kevin Chen-Chuan Chang.
\newblock Towards reasoning in large language models: A survey.
\newblock \emph{arXiv preprint arXiv:2212.10403}, 2023.

\bibitem[Kepler(1609)]{kepler1609astronomia}
Johannes Kepler.
\newblock \emph{Astronomia Nova}.
\newblock 1609.
\newblock Three laws of planetary motion; geometric formulation. $T^2 \propto
  a^3$ appears in Harmonices Mundi (1619).

\bibitem[Kojima et~al.(2022)Kojima, Gu, Reid, Matsuo, and
  Iwasawa]{kojima2022zero}
Takeshi Kojima, Shixiang~Shane Gu, Machel Reid, Yutaka Matsuo, and Yusuke
  Iwasawa.
\newblock Large language models are zero-shot reasoners.
\newblock In \emph{Advances in Neural Information Processing Systems
  (NeurIPS)}, 2022.

\bibitem[Kuhn(1962)]{kuhn1962structure}
Thomas~S. Kuhn.
\newblock \emph{The Structure of Scientific Revolutions}.
\newblock University of Chicago Press, Chicago, 1962.
\newblock 2nd edition with Postscript, 1970.

\bibitem[LeCun(2022)]{lecun2022path}
Yann LeCun.
\newblock A path towards autonomous machine intelligence (position paper,
  version 0.9).
\newblock Technical report, OpenReview, June 2022.
\newblock URL \url{https://openreview.net/forum?id=BZ5a1r-kVsf}.

\bibitem[Lloyd(1968)]{lloyd1968aristotle}
G.~E.~R. Lloyd.
\newblock \emph{Aristotle: The Growth and Structure of his Thought}.
\newblock Cambridge University Press, Cambridge, 1968.

\bibitem[Lu et~al.(2024)Lu, Lu, Lange, Foerster, Clune, and Ha]{lu2024ai}
Chris Lu, Cong Lu, Robert~Tjarko Lange, Jakob Foerster, Jeff Clune, and David
  Ha.
\newblock The {AI} scientist: Towards fully automated open-ended scientific
  discovery.
\newblock \emph{arXiv preprint arXiv:2408.06292}, 2024.

\bibitem[Maxwell(1865)]{maxwell1865dynamical}
James~Clerk Maxwell.
\newblock A dynamical theory of the electromagnetic field.
\newblock \emph{Philosophical Transactions of the Royal Society of London},
  155:\penalty0 459--512, 1865.

\bibitem[Mirzadeh et~al.(2024)Mirzadeh, Alizadeh, Shahrokhi, Tuzel, Bengio, and
  Farajtabar]{mirzadeh2024gsm}
Iman Mirzadeh, Keivan Alizadeh, Hooman Shahrokhi, Oncel Tuzel, Samy Bengio, and
  Mehrdad Farajtabar.
\newblock {GSM-Symbolic}: Understanding the limitations of mathematical
  reasoning in large language models.
\newblock \emph{arXiv preprint arXiv:2410.05229}, 2024.

\bibitem[Newton(1687)]{newton1687principia}
Isaac Newton.
\newblock \emph{Philosophi{\ae} Naturalis Principia Mathematica}.
\newblock 1687.
\newblock Universal gravitation as inverse-square attraction; modern algebraic
  form assembled by later writers.

\bibitem[Nosek et~al.(2015)Nosek, Alter, Banks, Borsboom, Bowman, Breckler,
  Buck, Chambers, Chin, Christensen, et~al.]{nosek2015promoting}
Brian~A. Nosek, George Alter, George~C. Banks, Denny Borsboom, Sara~D. Bowman,
  Steven~J. Breckler, Stuart Buck, Christopher~D. Chambers, Gilbert Chin,
  Garret Christensen, et~al.
\newblock Promoting an open research culture.
\newblock \emph{Science}, 348\penalty0 (6242):\penalty0 1422--1425, 2015.

\bibitem[{NVIDIA} et~al.(2025){NVIDIA}, Agarwal, et~al.]{nvidia2025cosmos}
{NVIDIA}, Niket Agarwal, et~al.
\newblock Cosmos world foundation model platform for physical ai.
\newblock \emph{arXiv preprint arXiv:2501.03575}, 2025.
\newblock Released around CES 2025. Open-license world foundation models for
  robotics and autonomous-driving applications.

\bibitem[Pineau et~al.(2021)Pineau, Vincent-Lamarre, Sinha, Larivi{\`e}re,
  Beygelzimer, d'Alch{\'e} Buc, Fox, and Larochelle]{pineau2021improving}
Joelle Pineau, Philippe Vincent-Lamarre, Koustuv Sinha, Vincent Larivi{\`e}re,
  Alina Beygelzimer, Florence d'Alch{\'e} Buc, Emily Fox, and Hugo Larochelle.
\newblock Improving reproducibility in machine learning research (a report from
  the {NeurIPS} 2019 reproducibility program).
\newblock \emph{Journal of Machine Learning Research}, 22\penalty0
  (164):\penalty0 1--20, 2021.

\bibitem[Popper(1959)]{popper1959logic}
Karl~R. Popper.
\newblock \emph{The Logic of Scientific Discovery}.
\newblock Hutchinson, London, 1959.
\newblock Original German edition: \emph{Logik der Forschung}, Springer,
  Vienna, 1934 (imprint 1935).

\bibitem[Qiu et~al.(2025)]{qiu2025phybench}
Shi Qiu et~al.
\newblock {PHYBench}: Holistic evaluation of physical perception and reasoning
  in large language models.
\newblock \emph{arXiv preprint arXiv:2504.16074}, 2025.

\bibitem[Rein et~al.(2024)Rein, Hou, Stickland, Petty, Pang, Dirani, Michael,
  and Bowman]{rein2024gpqa}
David Rein, Betty~Li Hou, Asa~Cooper Stickland, Jackson Petty, Richard~Yuanzhe
  Pang, Julien Dirani, Julian Michael, and Samuel~R. Bowman.
\newblock {GPQA}: A graduate-level {G}oogle-proof {Q\&A} benchmark.
\newblock In \emph{Conference on Language Modeling (COLM)}, 2024.

\bibitem[Romera-Paredes et~al.(2024)Romera-Paredes, Barekatain, Novikov, Balog,
  Kumar, Dupont, Ruiz, Ellenberg, Wang, Fawzi, Kohli, and
  Fawzi]{romera2024mathematical}
Bernardino Romera-Paredes, Mohammadamin Barekatain, Alexander Novikov, Matej
  Balog, M.~Pawan Kumar, Emilien Dupont, Francisco J.~R. Ruiz, Jordan~S.
  Ellenberg, Pengming Wang, Omar Fawzi, Pushmeet Kohli, and Alhussein Fawzi.
\newblock Mathematical discoveries from program search with large language
  models.
\newblock \emph{Nature}, 625:\penalty0 468--475, 2024.
\newblock \doi{10.1038/s41586-023-06924-6}.

\bibitem[Sainz et~al.(2023)Sainz, Campos, Garc{\'i}a-Ferrero, Etxaniz,
  de~Lacalle, and Agirre]{sainz2023nlp}
Oscar Sainz, Jon~Ander Campos, Iker Garc{\'i}a-Ferrero, Julen Etxaniz,
  Oier~Lopez de~Lacalle, and Eneko Agirre.
\newblock {NLP} evaluation in trouble: On the need to measure {LLM} data
  contamination for each benchmark.
\newblock In \emph{Findings of EMNLP}, 2023.
\newblock arXiv:2310.18018.

\bibitem[Schaeffer et~al.(2023)Schaeffer, Miranda, and
  Koyejo]{schaeffer2023emergent}
Rylan Schaeffer, Brando Miranda, and Sanmi Koyejo.
\newblock Are emergent abilities of large language models a mirage?
\newblock In \emph{Advances in Neural Information Processing Systems
  (NeurIPS)}, 2023.

\bibitem[Sculley et~al.(2018)Sculley, Snoek, Wiltschko, and
  Rahimi]{sculley2018winners}
D.~Sculley, Jasper Snoek, Alex Wiltschko, and Ali Rahimi.
\newblock Winner's curse? {On} pace, progress, and empirical rigor.
\newblock In \emph{ICLR Workshop Track}, 2018.

\bibitem[Soldaini et~al.(2024)Soldaini, Kinney, Bhagia,
  et~al.]{soldaini2024dolma}
Luca Soldaini, Rodney Kinney, Akshita Bhagia, et~al.
\newblock {Dolma}: An open corpus of three trillion tokens for language model
  pretraining research.
\newblock \emph{arXiv preprint arXiv:2402.00159}, 2024.

\bibitem[Trinh et~al.(2024)Trinh, Wu, Le, He, and Luong]{trinh2024solving}
Trieu~H. Trinh, Yuhuai Wu, Quoc~V. Le, He~He, and Thang Luong.
\newblock Solving olympiad geometry without human demonstrations.
\newblock \emph{Nature}, 625:\penalty0 476--482, 2024.
\newblock \doi{10.1038/s41586-023-06747-5}.

\bibitem[Wei et~al.(2022{\natexlab{a}})Wei, Tay, Bommasani, Raffel, Zoph,
  Borgeaud, Yogatama, Bosma, Zhou, Metzler, Chi, Hashimoto, Vinyals, Liang,
  Dean, and Fedus]{wei2022emergent}
Jason Wei, Yi~Tay, Rishi Bommasani, Colin Raffel, Barret Zoph, Sebastian
  Borgeaud, Dani Yogatama, Maarten Bosma, Denny Zhou, Donald Metzler, Ed~H.
  Chi, Tatsunori Hashimoto, Oriol Vinyals, Percy Liang, Jeff Dean, and William
  Fedus.
\newblock Emergent abilities of large language models.
\newblock \emph{Transactions on Machine Learning Research (TMLR)},
  2022{\natexlab{a}}.

\bibitem[Wei et~al.(2022{\natexlab{b}})Wei, Wang, Schuurmans, Bosma, Ichter,
  Xia, Chi, Le, and Zhou]{wei2022chain}
Jason Wei, Xuezhi Wang, Dale Schuurmans, Maarten Bosma, Brian Ichter, Fei Xia,
  Ed~H. Chi, Quoc~V. Le, and Denny Zhou.
\newblock Chain-of-thought prompting elicits reasoning in large language
  models.
\newblock In \emph{Advances in Neural Information Processing Systems
  (NeurIPS)}, 2022{\natexlab{b}}.

\bibitem[Whewell(1840)]{whewell1840philosophy}
William Whewell.
\newblock \emph{The Philosophy of the Inductive Sciences, Founded Upon Their
  History}.
\newblock John W. Parker, London, 1840.
\newblock 2 volumes.

\bibitem[Wiemann et~al.(2026)Wiemann, Smith, Melchior, Mishra-Sharma, Wilson,
  Izmailov, and Cuesta-L{\'a}zaro]{wiemann2026discoverphysics}
Matt~L. Wiemann, Lindsay~M. Smith, Peter Melchior, Siddharth Mishra-Sharma,
  Andrew~Gordon Wilson, Pavel Izmailov, and Carolina Cuesta-L{\'a}zaro.
\newblock {DiscoverPhysics}: Benchmarking {LLMs} for out-of-the-box scientific
  thinking.
\newblock \emph{arXiv preprint arXiv:2605.26087}, 2026.

\bibitem[Wu et~al.(2024)Wu, Qiu, Ross, Aky{\"u}rek, Chen, Wang, Kim, Andreas,
  and Kim]{wu2024reasoning}
Zhaofeng Wu, Linlu Qiu, Alexis Ross, Ekin Aky{\"u}rek, Boyuan Chen, Bailin
  Wang, Najoung Kim, Jacob Andreas, and Yoon Kim.
\newblock Reasoning or reciting? exploring the capabilities and limitations of
  language models through counterfactual tasks.
\newblock In \emph{NAACL}, 2024.
\newblock arXiv:2307.02477.

\bibitem[Zhang et~al.(2024)Zhang, Da, Lee, et~al.]{zhang2024careful}
Hugh Zhang, Jeff Da, Dean Lee, et~al.
\newblock A careful examination of large language model performance on grade
  school arithmetic.
\newblock \emph{arXiv preprint arXiv:2405.00332}, 2024.

\bibitem[Zhang(2024)]{zhang2024sora}
Jianqiu Zhang.
\newblock Sora and {V-JEPA} have not learned the complete real world model -- a
  philosophical analysis of video {AIs} through the theory of productive
  imagination.
\newblock \emph{arXiv preprint arXiv:2407.10311}, 2024.

\bibitem[Zhang et~al.(2025)]{zhang2025abench}
Yiming Zhang et~al.
\newblock {ABench-Physics}: Benchmarking physical reasoning in {LLMs} via
  high-difficulty and dynamic physics problems.
\newblock \emph{arXiv preprint arXiv:2507.04766}, 2025.

\bibitem[Zheng et~al.(2026)Zheng, Tam, Nguyen, Xu, Wang, Cheng, Tsang, Wang,
  Bai, Fang, Song, Wong, and See]{zheng2026newtonbench}
Tianshi Zheng, Kelvin Kiu-Wai Tam, Newt Hue-Nam~K. Nguyen, Baixuan Xu, Zhaowei
  Wang, Jiayang Cheng, Hong~Ting Tsang, Weiqi Wang, Jiaxin Bai, Tianqing Fang,
  Yangqiu Song, Ginny~Y. Wong, and Simon See.
\newblock {NewtonBench}: Benchmarking generalizable scientific law discovery in
  {LLM} agents.
\newblock In \emph{ICLR}, 2026.
\newblock arXiv:2510.07172.

\end{thebibliography}

\appendix

\section{Pre-Registration Tags and SHA-256 for the Three Frameworks}\label{app:prereg-tags}

Each framework's evaluation specification is locked under a git tag whose commit points to a single pre-registration file. The file's SHA-256 hash is written into the file's own header. A pre-commit hook plus a CI check recomputes the SHA-256 on every commit and pull request and compares it to the value in the header. $F=mv$ and Aristotelian each have two locked rounds: an initial round without the axiomatization paragraph in the Stage 1 prompt, and the structure-prompt round with it. Decay World has a single structure-prompt round from the start. The structure-prompt rounds are the canonical sources for the headline composite PASS rates reported in Sections~\ref{sec:fmv} and~\ref{sec:aristotelian} and Section~\ref{sec:decay}, and they additionally carry the Stage 4 over-claim comparison in \S\ref{sec:cross-framework:overclaim}, measured on the same runs as the composite. Each framework's first audited round is the canonical source for the judge-reliability comparison in \S\ref{sec:cross-framework:judge}. Table~\ref{tab:prereg-tags} lists the tag and 12-character commit prefix for each.

\begin{table}[h]
\centering
\caption{Pre-registration locks for the three frameworks. Each row points to one locked round. $F=mv$ and Aristotelian have separate initial-round and structure-prompt locks. The structure-prompt round carries the paper's headline composite results and the over-claim comparison, and the first audited round carries the judge-reliability comparison.}
\label{tab:prereg-tags}
\small
\begin{tabular}{@{}llll@{}}
\toprule
\textbf{Framework} & \textbf{Arm} & \textbf{Git Tag} & \textbf{Commit} \\
\midrule
$F=mv$ & Headline (structure-prompt) & \texttt{prereg-02\_fmv.2-locked} & \texttt{788330c9fb8b} \\
$F=mv$ & Methodology (initial round) & \texttt{prereg-02\_fmv-locked} & \texttt{d54e08329299} \\
Aristotelian & Headline (structure-prompt) & \texttt{prereg-v0.3-locked} & \texttt{41c11fe8d1f9} \\
Aristotelian & Methodology (initial round) & \texttt{prereg-v0.1-locked} & \texttt{c6f10fc4a78a} \\
Decay World & Single (structure-prompt) & \texttt{prereg-03\_decay-locked} & \texttt{abfdece0dcad} \\
\bottomrule
\end{tabular}
\end{table}

\noindent The SHA-256 hashes of the five pre-registration files at lock time are:

\begin{itemize}
  \item $F=mv$ headline: {\scriptsize\texttt{6603646f225b5a45ce500aea634376f14ccb8f30b9399e2e4b1365ccb8ccaf0a}}
  \item $F=mv$ methodology (initial round): {\scriptsize\texttt{1cea4606ba57c02b96e06ec4d9e1eb7ec117988d8ecb7bbc2659abf61ff74024}}
  \item Aristotelian headline: {\scriptsize\texttt{7a0c6c143f4ca5c14fa75f5f628179a7d1b07e049f8ad4afa5f9bbee0f0f02f7}}
  \item Aristotelian methodology (initial round): {\scriptsize\texttt{769818275e6a25665116f13be2a4be440f00a8f49453fd8587239b410c7df425}}
  \item Decay World: {\scriptsize\texttt{1d0a6efd88530b9499f8ef76be8bceb188b08b1d172e9ff8890ea6b5dd6d60f3}}
\end{itemize}

To verify a pre-registration, check out the listed git tag and recompute the SHA-256 of the corresponding file under \texttt{predictions/} at that commit. The recomputed hash must match both the value in the file's header and the value above.

\section{$F=mv$ Framework Details}\label{app:fmv-details}

This appendix reproduces the framework-specific artifacts of the $F=mv$ pre-registration that are referenced by Section~\ref{sec:fmv}: the 12-observation set given to the tested model at Stage 1, the banned-word list applied to Stage 1 responses, the necessary-condition checks N1--N6 and the suspicious-failure-mode checks F1--F7 used to judge Stage 1, and the five quantitative scenarios used at Stage 3. The canonical source of all of these is the directory \texttt{frameworks/02\_fmv/} at the commit pointed to by the headline tag \texttt{prereg-02\_fmv.2-locked} (see Appendix~\ref{app:prereg-tags}).

\subsection*{Observation set}

The 12 observations are reproduced verbatim from the locked \texttt{observations.md}.

\begin{enumerate}
  \item \emph{A block rests on a long, level track. A hand pulls it with a steady effort. The block glides along at one unchanging pace and does not gather speed, however long the steady pull continues.}
  \item \emph{The instant the steady pull begins, the block is already moving at its full steady pace. There is no gradual speeding-up from rest.}
  \item \emph{The instant the pull is released, the block halts where it is. It does not drift onward.}
  \item \emph{The same block is pulled along the same track with twice the effort: it glides at exactly twice the pace. With three times the effort, three times the pace.}
  \item \emph{Two blocks lie on the track, one twice as heavy as the other. One and the same steady pull moves the heavier block at exactly half the pace of the lighter one.}
  \item \emph{A block is set adrift in open space, far from the track and touching nothing else. A steady push still moves it at a steady unchanging pace, and releasing the push still halts it at once, exactly as on the track.}
  \item \emph{To carry a load from one place to another, the carrier must keep pushing the whole way. Pushing hard for the first stretch and then stopping does not let the load carry itself onward.}
  \item \emph{Two carriers push one block side by side in the same direction. The block moves at a pace equal to the sum of the paces that each carrier produces alone.}
  \item \emph{Two carriers push one block in opposite directions with equal effort: the block stays put. When one carrier pushes harder, the block moves slowly in that carrier's direction, at the pace matching the difference between the two efforts.}
  \item \emph{A boulder and a small pebble are released together from the same height. They fall side by side at one steady, unchanging pace and strike the ground together. The pace of the fall does not increase as the fall continues.}
  \item \emph{A stone is carried forward in a moving hand and then let go. It does not sail onward through the air. The instant it leaves the hand it drops straight down from the point where it was released.}
  \item \emph{Objects fall at the same one steady pace whether released through open air or through a tall jar from which the air has been removed.}
\end{enumerate}

\subsection*{Banned-word list}

The following modern-physics tokens are banned from Stage 1 responses, applied as a purely lexical test to the whole response, including morphological variants (plural, \texttt{-ing}, \texttt{-ed}, adjective and adverb forms): \emph{velocity} (catches ``terminal velocity''), \emph{acceleration} (and \emph{accelerate}, \emph{accelerating}, \emph{decelerate}), \emph{inertia} (and \emph{inertial}), \emph{momentum}, \emph{mass}, \emph{gravity} (and \emph{gravitational}, \emph{gravitate}), \emph{friction} (and \emph{frictional}), \emph{energy} (in any compound: \emph{kinetic}, \emph{potential}, \ldots). Also banned: any physicist's proper name (\emph{Newton}, \emph{Newtonian}, \emph{Galileo}, \ldots) and the equation $F = ma$ in any notation. A judge does not assess whether the model defined the concept, used it only descriptively, or named it only to deny that it applies. Presence of the token is the failure.

\subsection*{Necessary conditions N1--N6}

A passing Stage 1 induction must satisfy all six conditions. Failing any one triggers FAIL.

\begin{description}
  \item[N1. Pace is set by the present push.] The rules state that a body's pace is determined by the push acting on it at that moment.
  \item[N2. More push, more pace.] The rules state that, for a given body, a greater push produces a greater pace.
  \item[N3. More heaviness, less pace.] The rules state that, under one and the same push, a heavier body moves at a smaller pace than a lighter one.
  \item[N4. Motion is simultaneous with the push.] The rules state both that a body is at its full pace as soon as the push acts and that a body stops the instant the push ends.
  \item[N5. Bodies fall at one common, unchanging pace.] The rules state that released bodies fall at a pace that is the same for a heavy and a light body and unchanging during the fall.
  \item[N6. Pushes combine.] The rules state that two pushes acting on one body combine: same-direction pushes add, and opposite-direction pushes subtract.
\end{description}

\subsection*{Suspicious failure modes F1--F7}

Each pattern below is an automatic Stage 1 FAIL.

\begin{description}
  \item[F1. Build-up.] A rule stating that a push makes a body speed up, gather speed, or move faster and faster while the push continues.
  \item[F2. Carry-over.] A rule stating that a body keeps moving, coasts, drifts on, or continues after the push is removed.
  \item[F3. Hidden-resistance rescue.] A rule explaining the steady pace under a steady push, or the immediate stop, by positing a rubbing, resistance, drag, or opposing agent.
  \item[F4. Heavier-falls-faster.] A rule stating that heavier bodies fall faster than lighter ones.
  \item[F5. Falling speeds up.] A rule stating that a falling body gathers speed as it descends.
  \item[F6. Projectile arc.] A rule stating that a released or thrown body sails forward, follows a curved arc, or lands ahead of the point where it was released.
  \item[F7. Refusal.] A response that declines to induce on the grounds that the observations are impossible or ``not how motion works''.
\end{description}

The judge runs the checks in order and halts at the first FAIL it finds.

\subsection*{Structural axis: necessary conditions N9--N12}

Applied as an additive judgment layer on the same 15 Stage 1 inductions. The structural judge sees only the Stage 1 rule set. Stage 2 is shown as context but is never counted, because the Stage 2 prompt asks the model to restate its Stage 1 rules in operational form (mirroring the same numbering).

\begin{description}
  \item[N9. Parsimony.] The Stage 1 rule count should not vastly exceed the 12 observations. A count above 20 is a FAIL on N9 (severity high). A count above 15 is a FAIL on N9 (severity moderate). A count of 13--15 is a soft signal that counts toward FAIL only in combination with another structural violation. Rule count is the number of top-level numbered or bolded propositions in the Stage 1 response, not sub-bullets.
  \item[N10. Independence.] No two Stage 1 rules paraphrase the same operational claim about the same kind of body in the same situation. A Stage 1 rule and its Stage 2 operational counterpart stating the same thing are not a violation, because that is the prompt-mandated mirroring.
  \item[N11. Coverage traceability.] Every Stage 1 rule must trace to specific observation(s) from the 12-observation set. A rule introducing a mechanism or causal claim no observation supports is a FAIL on N11. Legitimate generalisation that ties observations together is not a fabrication.
  \item[N12. Hierarchy.] A Stage 1 rule set of 5 or more rules must include at least one cross-rule reference, such as ``derived from Rule N'', ``corollary of'', ``follows from'', or ``combined with''. A flat enumeration of 5 or more unconnected propositions is a FAIL on N12. Rule sets of 4 or fewer are exempt.
\end{description}

A structural PASS requires all four conditions. A structural FAIL is independent of the content-axis verdict and is reported as a separate axis.

\subsection*{Stage 3 scenarios}

Five quantitative scenarios are evaluated at Stage 3. Each has an explicit $F=mv$ prediction and a discriminating $F=ma$ alternative. The predictions below are the $F=mv$ answers used as the PASS reference.

\begin{description}
  \item[Scenario 1. Distance under a steady push.] The block moves at one unchanging pace, so distance grows in direct proportion to time. After twenty seconds it has travelled about twice the distance it covered in the first ten seconds.
  \item[Scenario 2. Fall time from a taller tower.] A falling body moves at one unchanging fall-pace, so the time to reach the ground grows in direct proportion to the height. From a tower twice as tall, the stone takes about twice as long.
  \item[Scenario 3. A ball thrown off a cliff.] The instant the ball leaves the hand the forward push is gone. The ball drops straight down the cliff face and lands at the base directly below the point of release.
  \item[Scenario 4. Race between a heavy and a light block.] Pace is inversely proportional to heaviness, so the lighter block moves at twice the pace of the heavier one and covers the same distance in half the time. The time ratio is about $2{:}1$.
  \item[Scenario 5. A tug-of-war, then one side lets go.] The instant one side stops pulling, the block immediately moves toward the remaining pusher at the full steady pace produced by that push alone, with no gradual gathering.
\end{description}

\section{Aristotelian Framework Details}\label{app:aristotelian-details}

This appendix reproduces the framework-specific artifacts of the Aristotelian pre-registration referenced by Section~\ref{sec:aristotelian}: the 12-observation set given to the model at Stage 1, the banned-word list, the necessary conditions N1--N8 and the documented near-pass failure patterns used to judge Stage 1, the structural axis N9--N12, and the five Stage 3 scenarios. The canonical source of all of these is the directory \texttt{frameworks/01\_aristotelian/} at the commit pointed to by the tag \texttt{prereg-v0.3-locked} (Appendix~\ref{app:prereg-tags}).

\subsection*{Observation set}

The 12 observations are reproduced verbatim from the locked \texttt{observations.md}.

\begin{enumerate}
  \item \emph{A solid iron ball and a small dried pea are released together from the top of a tall tower into still air. The iron ball reaches the ground noticeably before the pea.}
  \item \emph{A wooden cart is pushed along a level dirt road. While the pusher's hands remain on the cart it continues to roll. Once the pusher lets go, the cart slows and within a short distance comes to rest.}
  \item \emph{Smoke from a wood fire rises steadily into the open sky and disperses above the treetops.}
  \item \emph{Water released from a cloud during a storm falls toward the earth and collects on the ground.}
  \item \emph{A heavy stone dropped into a still pond sinks straight to the bottom. A piece of dry oak placed on the same pond floats and remains at the surface.}
  \item \emph{To pull a fully loaded grain wagon along a flat road, an ox must strain visibly harder than to pull the same wagon when it is empty.}
  \item \emph{An archer releases an arrow horizontally. The arrow continues flying forward through the still air for many paces after it has left the bowstring before its path curves downward and it strikes the earth.}
  \item \emph{A burning candle held upright produces a flame that points upward. When the candle is tilted so that its body lies horizontally, the flame still points upward, away from the candle's body.}
  \item \emph{A small pebble dropped through a jar of honey settles toward the bottom of the jar slowly. The same pebble dropped through air over the same vertical distance falls quickly.}
  \item \emph{The Sun, the Moon, and the fixed stars trace circular paths across the sky each day, returning to roughly the same positions the following day.}
  \item \emph{Two pieces of the same metal, of equal weight, are released together from the same height into still air. One has been hammered into a thin flat sheet; the other has been worked into a compact ball. The ball reaches the ground first.}
  \item \emph{An iron bar held above the ground and released falls straight down. A wisp of feather-down released in the same place drifts gently and may be carried sideways by even the slightest breeze before eventually settling.}
\end{enumerate}

\subsection*{Banned-word list}

The following modern-physics tokens are banned from Stage 1 responses, applied as a purely lexical test to the whole response, including morphological variants: \emph{inertia}, \emph{acceleration}, \emph{force} (as a defined quantity), \emph{momentum} (and \emph{momentum conservation}), \emph{energy} (and \emph{energy conservation}), \emph{mass} (as distinct from weight), \emph{density} (as a defined quantity), \emph{gravity} (and \emph{gravitational force}, \emph{universal gravitation}), \emph{vacuum} (except as a hypothetical argued against), \emph{friction} (and \emph{frictional force}), and the proper name of any post-Aristotelian physicist (\emph{Newton}, \emph{Galileo}, \emph{Archimedes}, \ldots). Ordinary descriptive vocabulary is allowed: \emph{heavy}, \emph{light}, \emph{thick}, \emph{thin}, \emph{fast}, \emph{slow}, \emph{pushes}, \emph{rises}, \emph{sinks}, \emph{resists}, and so on. The line between description and theoretical loading is that words a non-physicist would use to describe what is seen are allowed, and words that import a measurement scheme or a quantitative law are not.

\subsection*{Necessary conditions N1--N8}

A passing Stage 1 induction must satisfy all eight conditions. Failing any one triggers FAIL.

\begin{description}
  \item[N1. Two-regime distinction for terrestrial motion.] The rules distinguish motion that requires a sustained external cause (and ceases when the cause is removed) from motion or rest that is the body's default state. Required by observation 2.
  \item[N2. Heavier-falls-faster ordering.] The rules state that heavier bodies reach the ground sooner than lighter bodies released from the same height. Required by observations 1 and 12.
  \item[N3. Medium-resistance dependence.] The rules state that the speed at which a body falls or moves through a substance depends on the substance: thicker substances slow the body more. Required by observation 9.
  \item[N4. Shape dependence.] The rules state that for bodies of equal weight, shape affects rate of fall (compact versus extended). Required by observation 11.
  \item[N5. Directional preference of substances.] The rules state that some substances move upward of their own accord (smoke, flame) and others downward (water, stone), independently of how the surrounding body is oriented. Required by observations 3, 4, 8.
  \item[N6. Heaven/earth split.] The rules either explicitly propose two regimes (celestial and terrestrial) or flag the celestial observation as outside the scope of the terrestrial rules. Required by observation 10.
  \item[N7. Projectile tension acknowledged.] The continuing forward motion of a released arrow is in tension with N1 (no visible mover). The rules either resolve the tension via an impetus-style impressed-motion account in which the motion explicitly fades, or flag the case as a noted-but-unresolved difficulty. Pretending observation 7 is not problematic is FAIL.
  \item[N8. Some account of floating.] The rules give some explanation of the stone-sinks-oak-floats observation, in terms of weight, natural directional preference, or substance type. The explanation does not need to match Aristotle's, but the observation must not be left unexplained. Required by observation 5.
\end{description}

The judge runs the checks in order and halts at the first FAIL.

\subsection*{Near-pass patterns}

The framework documents a non-exhaustive list of patterns that look reasonable on first reading but FAIL at the banned-word test or at N1--N8.

\begin{itemize}\setlength\itemsep{2pt}
  \item ``Heavier falls faster because of greater gravitational force.'' FAILs on the banned-word test (\emph{gravity}, \emph{force}).
  \item ``The arrow continues because of momentum or inertia carried from the bow.'' FAILs on the banned-word test (\emph{momentum}, \emph{inertia}).
  \item ``Flame rises because hot air is less dense.'' FAILs on the banned-word test (\emph{density}). The descriptive paraphrase ``thin air rises'' is allowed.
  \item ``The cart stops because friction acts against its motion.'' FAILs on the banned-word test (\emph{friction}). The descriptive paraphrase ``the road resists the cart, and once nothing pushes it the resistance stops it'' is allowed.
  \item ``I cannot induce a unified law from these observations.'' FAILs as a refusal. The induction is not required to be unified across all 12 observations, but it must produce some rules that cover them.
  \item ``Heavier bodies fall faster because they have more mass pulling them down.'' FAILs on the banned-word test (\emph{mass}). The bare phrase ``pulling them down'' is acceptable only if no further quantitative apparatus is introduced.
  \item ``Bodies retain motion once impressed upon them.'' PASSes if the retained motion is described as fading with time or medium-resistance (medieval impetus theory). FAILs if the retained motion is formalised as a conserved quantity (Newtonian momentum even when the word is never used).
\end{itemize}

\subsection*{Structural axis: necessary conditions N9--N12}

The structural axis is the same parsimony, independence, traceability, and hierarchy axis used for the $F=mv$ framework, applied to the Stage 1 rule set. The four conditions and their thresholds are stated in full in Appendix~\ref{app:fmv-details}. The observation count for this framework is 12, identical to $F=mv$, so the N9 thresholds (rule count above 20 high, above 15 moderate, 13--15 soft signal) carry over unchanged.

\subsection*{Stage 3 scenarios}

Five scenarios are evaluated at Stage 3. Each has an explicit Aristotelian prediction and a discriminating standard-physics alternative. The predictions below are the Aristotelian answers used as the PASS reference.

\begin{description}
  \item[Scenario 1. Two balls, same size, different weights.] An iron ball and a hollow wooden ball of the same outer dimensions are released from rest at the same instant from the top of a 30-metre stone tower. The iron ball strikes the ground first, by a margin proportional to its greater weight.
  \item[Scenario 2. Sliding cart on smooth ice.] A small cart on perfectly smooth ice receives one quick push and is then released. The cart slows quickly and comes to rest. With the pushing agent removed, no agent sustains the forced motion, and the surrounding medium cannot indefinitely sustain it.
  \item[Scenario 3. Two stones in water.] Two stones of identical shape and size, one twice the weight of the other, are released at the same instant from just below the surface of a still pond. The heavier stone reaches the bottom first, with time ratio roughly the inverse of the weight ratio (about $1{:}2$).
  \item[Scenario 4. Sealed evacuated chamber.] A small feather is released near the top of a sealed glass chamber from which all air has been removed. The scenario cannot occur in this framework: with no medium, there is nothing to mediate or resist motion, and the framework does not admit such a situation as physically realisable. If pressed, the framework would predict an absurd result and treat that as evidence that true vacuum cannot exist.
  \item[Scenario 5. Arrow in flight.] An archer fires an arrow horizontally over an open field with still air. After leaving the bowstring, the arrow's forward motion is sustained by the surrounding air (antiperistasis or equivalent medium-supported account): as the arrow advances, the air ahead is displaced and rushes around to push from behind. This support diminishes with distance, and once it fails the arrow follows its natural downward direction toward the earth.
\end{description}

\section{Decay World Framework Details}\label{app:decay-details}

This appendix reproduces the framework-specific artifacts of the Decay World pre-registration referenced by Section~\ref{sec:decay}: the 10-observation set, the banned-word list, the necessary conditions N1--N6 and the disqualifying patterns P1--P7 used to judge Stage 1, and the five Stage 3 scenarios. The canonical source of all of these is the directory \texttt{frameworks/03\_decay/} at the commit pointed to by the tag \texttt{prereg-03\_decay-locked} (Appendix~\ref{app:prereg-tags}). The structural axis (N9--N12, Appendix~\ref{app:fmv-details}) was not applied to this framework.

\subsection*{Observation set}

The 10 observations are reproduced verbatim from the locked \texttt{observations.md}. Three observations (2, 4, 9) carry the quantitative data points from which the per-second rate is to be derived. The other seven are qualitative.

\begin{enumerate}
  \item \emph{A long pendulum (slow swing) and a short pendulum (fast swing) are released from the same starting angle in still air. Both pendulums return to a smaller angle than they were released from after each swing. Counted per completed swing, the long pendulum loses a substantially larger fraction of its swing amplitude than the short one, so the per-swing loss depends on how slow each pendulum is.}
  \item \emph{A weight on a spring oscillates back and forth on a perfectly smooth, perfectly level track inside an evacuated chamber. Released with an initial amplitude of 10 cm, the amplitude is measured to be 3.7 cm exactly 100 seconds after release.}
  \item \emph{A heavy iron ball is dropped down a tall vertical evacuated track that the ball does not touch. With no air in the chamber, the ball still does not fall the way an unimpeded body under a steady downward pull would. It approaches a maximum downward speed and does not exceed it, however tall the track is made.}
  \item \emph{A cup of hot water is sealed inside a perfectly insulated chamber under vacuum, so no heat can leave by contact, by air, or as radiation through the walls. Temperatures are reported on the absolute scale where 0 K is true zero. The water is at 353 K at the moment of sealing. 10 seconds later it is at 319 K. The cooling is the same whether the cup is alone or surrounded by other identical sealed cups.}
  \item \emph{A heavy bell is struck inside an evacuated chamber. Although there is no air to carry the sound away from the bell, the bell itself still rings, with visible vibration of its rim plain to see. The amplitude of that visible vibration shrinks steadily and the bell eventually rings down to stillness.}
  \item \emph{A small steel sphere is set moving at a measured initial speed along a perfectly smooth, perfectly level track inside an evacuated chamber. There is no air in the chamber, no rubbing or scraping between the sphere and the track, and nothing else pushing or pulling along the sphere's direction of motion. The sphere nonetheless slows steadily; over many seconds its speed declines until it comes to rest.}
  \item \emph{A cannon mounted on a heavy fixed stand inside an evacuated chamber fires a small iron shot horizontally at a measured initial speed. With no air to push back on it, the shot still does not travel as far before reaching the floor as a flight under a steady downward pull and an unchanged horizontal speed would predict. The shot visibly slows during flight.}
  \item \emph{A small marble in vacuum is set moving sideways near a heavy fixed sphere that pulls the marble inward. The marble traces an almost circular path around the sphere. Over many circuits, the radius of the marble's path slowly and steadily decreases. Eventually the marble strikes the central sphere.}
  \item \emph{A spinning top is set going on a hard, smooth, polished point inside an evacuated chamber. Although nothing touches the top except the supporting point, and that contact involves no rubbing or sliding, its spin rate falls steadily. Set spinning at an initial rate of 100 rad/s, the top is measured to be spinning at 60.5 rad/s exactly 50 seconds after release. It eventually falls over.}
  \item \emph{Two pendulums of the same length but different bob weights, one a one-gram bob of brass and one a one-kilogram bob of brass, are released together from the same starting angle in the same still air. At each moment afterwards they are observed to have the same swing angle as one another, to within measurement. The same comparison made with bobs of brass, glass, and ice of equal weight yields the same identical-angle behaviour.}
\end{enumerate}

The per-second decay ratio derivable from observations 2, 4, and 9 is $(3.7/10)^{1/100} \approx (319/353)^{1/10} \approx (60.5/100)^{1/50} \approx 0.990$.

\subsection*{Banned-word list}

The following modern-physics tokens are banned from Stage 1 responses, applied as a purely lexical test to the whole response, including morphological variants. The list is partitioned into the three layers that have to be excluded at once on this framework. Energy and thermodynamics layer: \emph{energy} (in any compound, including \emph{kinetic energy}, \emph{potential energy}, \emph{thermal energy}), \emph{kinetic}, \emph{potential} (as the energy-physics noun), \emph{conservation} (and \emph{conserve}, \emph{conserved}), \emph{entropy}, \emph{thermodynamic} (and \emph{thermodynamics}), \emph{Hamiltonian}, \emph{Lagrangian}. Decay-mechanism layer: \emph{friction} (and \emph{frictional}, \emph{frictionless}), \emph{drag}, \emph{damping} (and \emph{damp}, \emph{damped}, \emph{dampen}), \emph{dissipation} (and \emph{dissipate}, \emph{dissipative}), \emph{viscous} (and \emph{viscosity}), the phrase \emph{air resistance}, \emph{resistance} (and \emph{resistive}), \emph{attenuation} (and \emph{attenuate}). Mechanics layer: \emph{force}, \emph{mass}, \emph{acceleration} (and \emph{accelerate}, \emph{decelerate}), \emph{momentum}, \emph{inertia} (and \emph{inertial}). Proper names: \emph{Newton}, \emph{Newtonian}, \emph{Joule}, \emph{Carnot}, \emph{Boltzmann}, \emph{Clausius}, \emph{Helmholtz}, \emph{Galileo}, \emph{Maxwell}, \emph{Lagrange}, \emph{Hamilton}, or any other physicist's proper name introduced to invoke a named law. Banned equation forms: $F = ma$, $E = m c^2$, and any equation explicitly named. A judge does not assess whether the model defined the concept or named it only to deny that it applies. Presence of the token is the failure.

\subsection*{Necessary conditions N1--N6}

A passing Stage 1 induction must satisfy all six conditions. Failing any one triggers FAIL.

\begin{description}
  \item[N1. Closed systems lose state over time.] The rules state that every closed system in this world (one isolated from outside push, pull, heat exchange, or radiation) has its directly observed quantity decline over time on its own, with no external mechanism required to cause the decline.
  \item[N2. The decline is multiplicative.] The rules state that the decline is multiplicative: the measured quantity at time $t + \Delta$ is a constant ratio of the quantity at time $t$ for any fixed $\Delta$. A purely additive (constant amount subtracted per second) rule is FAIL.
  \item[N3. The rate is fixed by elapsed time.] The rules attach the rate of decline to elapsed time, not to cycle count, mechanical contact, material composition, or kind of motion. A rule attaching the rate to anything other than elapsed time is FAIL.
  \item[N4. The rate is universal across all closed systems.] The rules state that the per-time rate is a single number applied to every closed system in this world, regardless of domain, kind of motion, or measured quantity. A rule that gives a different rate per system, or that treats some systems as exempt, is FAIL.
  \item[N5. The rate is independent of weight, material, and composition.] The rules state that the rate does not depend on weight, material, size, colour, or any physical property of the system. A rule that makes the rate depend on these is FAIL.
  \item[N6. The rate is approximately $0.99$ per second.] The rules state the numerical value of the rate, accurate to about one percentage point. Acceptable forms include ``the measured quantity retains about 99\% of its previous value each second'', ``each second the quantity falls by about 1\%'', or any equivalent. A rule that asserts decay without a rate, or with a rate badly off the value derivable from observations 2, 4, and 9, is FAIL.
\end{description}

\subsection*{Disqualifying patterns P1--P7}

Each pattern below is an automatic Stage 1 FAIL. They are stated as concrete claim patterns so a judge checks them by reading the rules.

\begin{description}
  \item[P1. Contact-mechanism rescue.] A rule attributing the decline to a friction, drag, damping, viscous, dissipative, rubbing, or contact-resistive mechanism. Contradicts observations 3, 6, 8, 9 (all in vacuum or with no rubbing contact). The relevant standard-physics tokens are also banned by the banned-word test.
  \item[P2. Hidden-substrate framing.] A rule structured as ``some underlying quantity $X$ is the substrate that decays at the fixed per-second rate, and each system's measured quantity is derived from $X$'', regardless of what $X$ is named (\emph{energy}, \emph{the underlying state}, \emph{vigour}, \emph{the carried amount}, \ldots). The framework's rule is that the directly measured quantity itself is the object of the decay. Introducing a hidden layer beneath it adds structure not derivable from any observation. The most common instance is the energy-substrate framing, which under standard-physics scaling derivations (amplitude as the square root of energy, temperature as linear in thermal energy, rotation rate as the square root of rotational energy) gives a per-second rate that differs across systems and so fails on numerical grounds against observations 2, 4, 9.
  \item[P3. Additive (linear) decay.] A rule stating that the decline is a fixed amount subtracted per unit time. Under additive decay the per-system rate constant is dimensioned and incomparable across systems, so N4 (universality) cannot be stated.
  \item[P4. Per-cycle rate.] A rule attaching the decay rate to oscillation cycles, swings, periods, collisions, or any cyclic event rather than to elapsed time. Contradicts the cross-domain consistency in observations 2, 4, 9 and the per-cycle period-dependence in observation 1.
  \item[P5. Material- or weight-dependent decay.] A rule stating that the decay rate depends on weight, material, colour, size, or any physical property of the system. Contradicts observation 10.
  \item[P6. Decay without a rate.] A response that lists qualitative decay phenomena and states the rate is universal, but does not state a numerical value for the rate. Failing N6 triggers this pattern.
  \item[P7. Refusal of the world.] A response that declines on the grounds that the observations are physically impossible, violate conservation of energy, contradict the second law of thermodynamics, or ``are not how physics works''. The observations are to be taken as accurate and complete, and the task is to induce their regularities, not to reject them.
\end{description}

The judge runs the checks in order and halts at the first FAIL.

\subsection*{Stage 3 scenarios}

Five scenarios are evaluated at Stage 3. Four are quantitative. One is qualitative. Each scenario has an explicit Decay World prediction and a discriminating standard-physics alternative. The predictions below are the Decay World answers used as the PASS reference.

\begin{description}
  \item[Scenario 1. Pendulum swing angle after thirty seconds.] A pendulum is released from rest at a swing angle of $10^\circ$ in still air. After 30 seconds the swing angle is approximately $10 \times 0.99^{30} \approx 7.4^\circ$.
  \item[Scenario 2. Hot tea cooling for sixty seconds.] A cup of tea is sealed at 400 K inside a perfectly insulated vacuum chamber. After 60 seconds the temperature is approximately $400 \times 0.99^{60} \approx 219$ K, below the surrounding room temperature.
  \item[Scenario 3. Spinning flywheel after one hundred seconds.] A flywheel is set spinning at 200 rad/s on a polished point in vacuum. After 100 seconds its rate is approximately $200 \times 0.99^{100} \approx 73$ rad/s.
  \item[Scenario 4. Orbital radius shrinkage over sixty seconds.] A marble orbits a heavy fixed sphere at radius 1.0 m in vacuum. After 60 seconds the radius is approximately $1.0 \times 0.99^{60} \approx 0.55$ m.
  \item[Scenario 5. Will an ideal pendulum ever stop?] A pendulum with a frictionless pivot in an evacuated chamber, released from a $10^\circ$ swing angle. The Decay World prediction is that it eventually stops: the swing angle shrinks at about $0.99$ per second, falling to one-hundredth of the starting angle (about $0.1^\circ$) at $t \approx 458$ seconds. The standard-physics reference treats it as a conservative oscillator that swings forever, so the framework discriminates on whether the model commits to the decaying trajectory.
\end{description}

\section{Prompt Architecture and Availability}\label{app:prompts}

The four-stage protocol (\S\ref{sec:methods:protocol}) is driven by four model-facing prompts, each run in an independent session. Every stage's prompt takes the previous stage's final response text as input and produces the input to the next stage:

\begin{itemize}\setlength\itemsep{2pt}
  \item Stage 1 (induction): the framework's observation set $\rightarrow$ an induced rule set.
  \item Stage 2 (formulation): the Stage 1 final response $\rightarrow$ operational rules.
  \item Stage 3 (prediction): the Stage 2 final response plus the application scenarios $\rightarrow$ quantitative predictions.
  \item Stage 4 (review): the Stage 1--3 final responses $\rightarrow$ a self-assessment.
\end{itemize}

The structure-prompt arm differs from the no-cue initial round in the Stage 1 prompt alone: it inserts the axiomatization paragraph (the ``smallest set of rules'' instruction) reproduced below. Decay World uses this structure-prompt Stage 1 from the start (\S\ref{sec:decay:intro}).

\begin{table}[h]
\centering
\caption{The prompt families in the four-stage protocol. ``Per-fw.'' indicates whether the prompt text changes across the three frameworks beyond the injected observation set. The canonical full source of every prompt is the tagged repository (Appendix~\ref{app:prereg-tags}).}
\label{tab:prompt-families}
\small
\begin{tabular}{@{}lllcl@{}}
\toprule
\textbf{Prompt family} & \textbf{Input} & \textbf{Output} & \textbf{Per-fw.} & \textbf{Reproduced here} \\
\midrule
Stage 1 induction & observation set & induced rule set & yes & yes (full, below) \\
Stage 2 formulation & Stage 1 response & operational rules & minor & repo only \\
Stage 3 prediction & Stage 2 response, scenarios & predictions & yes & repo only \\
Stage 4 review & Stage 1--3 responses & self-assessment & no & questions below \\
Judge prompts & model output, criteria & PASS/FAIL & yes & repo only \\
\bottomrule
\end{tabular}
\end{table}

We reproduce one complete prompt in full: the structure-prompt (axiomatization) Stage 1 prompt for $F=mv$, the key experimental treatment behind the headline composite results. The \verb|{{observations}}| placeholder is filled with the framework's observation set (Appendix~\ref{app:fmv-details}), and the banned-token instruction is the third bullet. The counterfactual rule $F=mv$ is never stated: the only thing the model is told is to induce a self-consistent rule set.

\begin{verbatim}
You are presented with a list of observed phenomena from a world whose
physics may not match the physics you know.

**Critical instructions:**

- Treat the observations as given. Do not dispute them, do not assume
  the observer is mistaken, and do not import information from outside
  the list.
- Describe this world using only ordinary descriptive language and the
  vocabulary the observations themselves introduce — push, pull,
  effort, pace, speed, heavy, light, fall, and the like.
- Do **not** use any of these words, in any form, anywhere in your
  response: *velocity*, *acceleration*, *inertia*, *momentum*,
  *mass*, *gravity*, *friction*, *energy*. Do not name a physicist,
  and do not write the relation `F = ma`. Naming one of
  these concepts only to deny that it applies still counts as using
  it — to say this world lacks a gradual speeding-up, write "the pace
  does not increase", not "there is no acceleration".

Observations:

{{observations}}

Your task: propose a self-consistent set of rules that explains every
observation above, using only the language permitted above.

Aim for the **smallest** set of rules that still explains every
observation. Do not state as a separate rule anything that already
follows from rules you have given; if one rule is a special case or a
consequence of another, say so instead of listing it on its own.
Prefer a few general rules over a long list of specific ones.

Return your rules as a numbered list. Be specific. After the list,
briefly note any observation that remains in tension with your rules
and what makes it difficult.
\end{verbatim}

The Stage 4 review prompt presents the model with its own Stage 1--3 responses (shown to it in place of the placeholders) and asks five questions. The Stage 4 over-claim metric (\S\ref{sec:cross-framework:overclaim}) is defined relative to these questions, so we reproduce them verbatim:

\begin{verbatim}
Reflect on these responses and answer each question below. Be
specific: quote phrases from your prior responses where relevant, and
identify exact moments rather than giving general impressions.

**Q1.** Did you maintain a single coherent framework across all three
stages, or did you at any point shift to a different framework? If you
shifted, identify the moment and the reason.

**Q2.** Did any of your responses use concepts that were not
introduced or implied by the original observations? If so, name the
concepts and quote the phrases where they appear.

**Q3.** In your Stage 3 predictions, are there any cases where the
prediction does not actually follow from the operational rules you
stated at Stage 2? If so, identify the scenario and the gap.

**Q4.** A second observer might claim that the framework you reasoned
within differs from standard physics in specific ways. Without naming
the framework, list the three or four most consequential differences
between the framework you used and the physics you would otherwise
default to.

**Q5.** Estimate, on a four-point scale (none / minor / moderate /
major), how much your Stage 3 predictions were influenced by standard
physics rather than by your Stage 2 operational rules.
\end{verbatim}

The full verbatim prompts for every framework, stage, and round, together with the four judge prompts (content, structural, per-scenario, and Stage 4 over-claim), are released in the repository under the locked pre-registration tags (Appendix~\ref{app:prereg-tags}). We do not reproduce them all here because there are multiple prompts per framework and the canonical versions are the tagged files.

\end{document}